\pdfoutput=1

\documentclass[11pt]{article}

\usepackage[preprint]{acl}
\usepackage{times}
\usepackage{latexsym}

\usepackage[T1]{fontenc}

\usepackage[utf8]{inputenc}

\usepackage{microtype}

\usepackage{inconsolata}

\usepackage{graphicx}
\usepackage{booktabs}
\usepackage{amsmath}
\usepackage{amssymb}
\usepackage{multirow}
\usepackage{chemformula}
\usepackage{xcolor}
\usepackage{pifont}
\usepackage{xspace}

\newcommand{\git}{\raisebox{-1.5pt}{\includegraphics[height=1.05em]{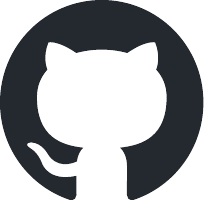}}\xspace}
\newcommand{\hf}{\raisebox{-1.5pt}{\includegraphics[height=1.05em]{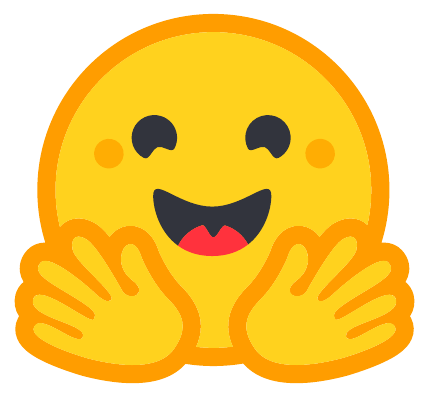}}\xspace}

\newcommand{\cmark}{\textcolor{green!60!black}{\ding{51}}}  
\newcommand{\xmark}{\textcolor{red}{\ding{55}}}            

\usepackage{colortbl}
\usepackage{multirow}

\definecolor{color1}{RGB}{255,99,71}   
\definecolor{color2}{RGB}{255,140,0}   
\definecolor{color3}{RGB}{255,255,102} 
\definecolor{color4}{RGB}{88,214,141}  
\definecolor{color5}{RGB}{50,130,210}  
\definecolor{color6}{RGB}{102,153,255} 
\definecolor{grayColor}{RGB}{211,211,211}

\title{FinRAGBench-V: A Benchmark for Multimodal RAG with Visual Citation in the Financial Domain}

\author{
 \textbf{Suifeng Zhao\textsuperscript{1}},
 \textbf{Zhuoran Jin\textsuperscript{2}},
 \textbf{Sujian Li\textsuperscript{3*}},
 \textbf{Jun Gao\textsuperscript{1}\thanks{Corresponding authors}}\\
 \textsuperscript{1}Key Laboratory of High Confidence Software Technologies, CS, Peking University, China\\
 \textsuperscript{2}School of Artificial Intelligence, University of Chinese Academy of Sciences\\
 \textsuperscript{3}State Key Laboratory of Multimedia Information Processing, School of Computer Sciences, Peking University
 \\
 \texttt{sfzhao25@stu.pku.edu.cn},
 \texttt{zhuoran.jin@nlpr.ia.ac.cn},
\texttt{\{lisujian,gaojun\}@pku.edu.cn}\\
\hf Dataset: \url{https://huggingface.co/datasets/zhaosuifeng/FinRAGBench-V}  \\
\git Code: \url{https://github.com/zhaosuifeng/FinRAGBench-V} 
}

\begin{document}
\maketitle

\begin{abstract}
Retrieval-Augmented Generation (RAG) plays a vital role in the financial domain, powering applications such as real-time market analysis, trend forecasting, and interest rate computation. However, most existing RAG research in finance focuses predominantly on textual data, overlooking the rich visual content in financial documents, resulting in the loss of key analytical insights. To bridge this gap, we present \textbf{FinRAGBench-V}, a comprehensive visual RAG benchmark tailored for finance which effectively integrates multimodal data and provides visual citation to ensure traceability. It includes a bilingual retrieval corpus with 60,780 Chinese and 51,219 English pages, along with a high-quality, human-annotated question-answering (QA) dataset spanning heterogeneous data types and seven question categories. Moreover, we introduce \textbf{RGenCite}, an RAG baseline that seamlessly integrates visual citation with generation. Furthermore, we propose an \textbf{automatic citation evaluation method} to systematically assess the visual citation capabilities of Multimodal Large Language Models (MLLMs). Extensive experiments on RGenCite underscore the challenging nature of FinRAGBench-V, providing valuable insights for the development of multimodal RAG systems in finance.

\end{abstract}

\section{Introduction}
Retrieval-Augmented Generation (RAG) \citep{DBLP:journals/jmlr/IzacardLLHPSDJRG23,DBLP:conf/icml/GuuLTPC20,DBLP:conf/nips/YuP0WYZSC24} has become a crucial approach for enhancing the performance of Large Language Models (LLMs) by integrating external knowledge with internal knowledge \cite{DBLP:conf/nips/YangSXSBCCGJJKM24,DBLP:conf/emnlp/Han00XWLWMC24,DBLP:journals/corr/abs-2403-10131}. This approach has been applied in a wide range of domain-specific tasks, among which, the financial domain is particularly representative due to its heavy reliance on complex multimodal data, such as line charts showing price fluctuations and tables presenting financial statistics. Therefore, it is critical to build a multimodal RAG system tailored to finance to enable reliable, explainable, and data-grounded analysis.

\begin{figure*}[t]
  \centering \includegraphics[clip=true,width=1\textwidth]{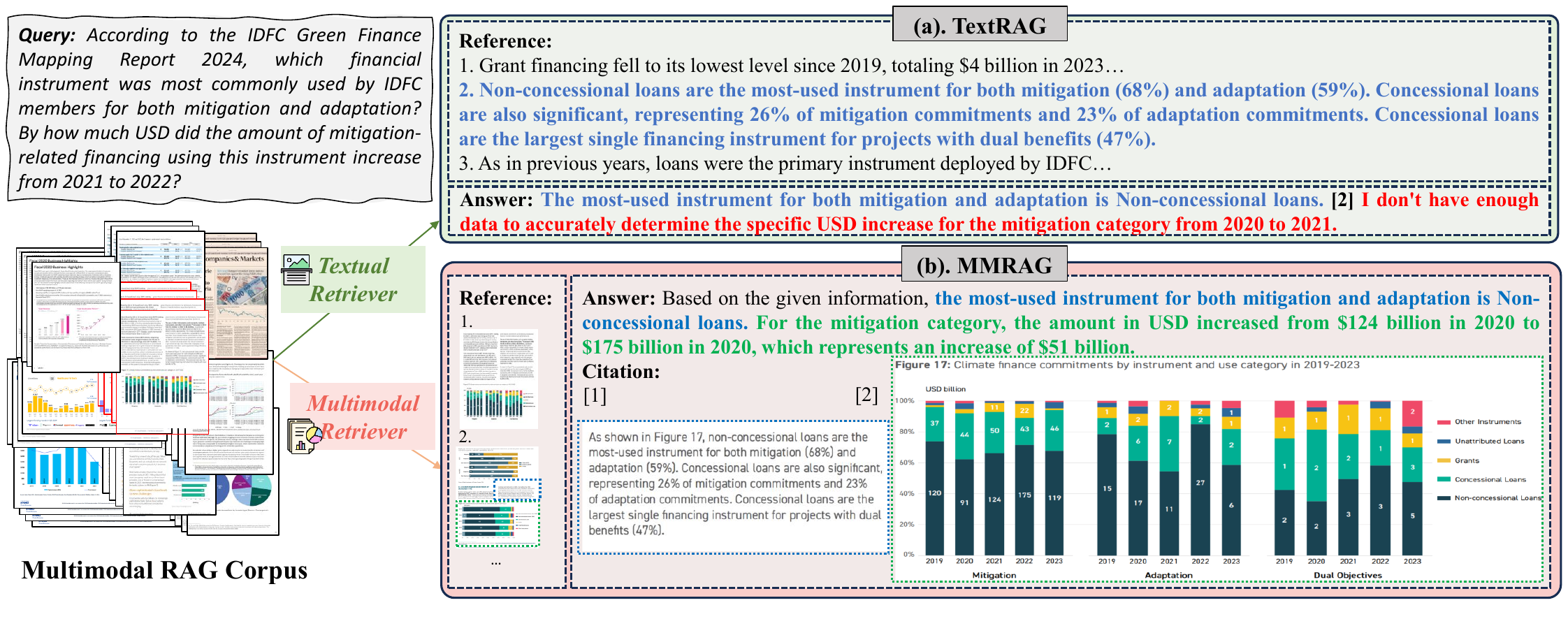}
  \caption{An example of a financial question requiring both text and visual understanding. (a) shows text-only RAG, where information loss leads to insufficient data for the model to answer the question. (b) illustrates our proposed paradigm, in which the model not only answers correctly based on retrieved information but also provides appropriate visual citations.
  }
  \label{fig:intro}
\end{figure*}

However, existing financial RAG efforts, such as FinQA \cite{DBLP:conf/emnlp/ChenCSSBLMBHRW21} and OmniEval \cite{DBLP:journals/corr/abs-2412-13018}, predominantly focus on text-only RAG, which may lose critical information when converting multimodal documents into plain text. As a result, they frequently fail to answer questions accurately, as shown in Figure \ref{fig:intro} (a). Although MME-Finance \citep{DBLP:journals/corr/abs-2411-03314} introduces a multimodal reasoning benchmark, it relies mostly on isolated screenshots and lacks retrieval support. Consequently, it falls short of reflecting the complexity of real-world financial scenarios, where answering questions often requires diverse data sources and heterogeneous data types. Furthermore, given the critical importance of precision in finance, RAG systems must ensure not only accuracy responses but also their traceability and verifiability, yet most existing benchmarks overlook these needs. Thus, designing a more comprehensive benchmark for multimodal RAG in finance is imperative. 

In this work, we propose \textbf{FinRAGBench-V}, a multimodal RAG benchmark tailored for finance, featuring grounded visual citation. This benchmark effectively integrates multimodal data and provides visual citations to ensure traceability, as shown in Figure \ref{fig:intro} (b). Specifically, we construct a large-scale retrieval corpus from diverse real-world financial sources, comprising 60,780 Chinese pages from 1,104 documents and 51,219 English pages from 1,105 documents, including research reports, financial statements, prospectuses, etc. In addition, we develop a high-quality financial question-answering (QA) dataset using GPT-4o \cite{DBLP:journals/corr/abs-2303-08774} assistance with manual verification. The dataset consists of 855 Chinese and 539 English QA pairs, covering a wide range of distinctive financial tasks, with questions categorized by data heterogeneity, including text, charts, and tables, and reasoning type, such as time-sensitive reasoning, numerical calculations, multi-page reasoning, etc. 

Based on this benchmark, we propose \textbf{RGenCite}, a simple yet effective multimodal RAG baseline that integrates retrieval, generation, and visual citation in a unified pipeline. The model is tasked with not only generating answers from retrieved contexts but also performing visual citation towards relevant document pages and specific content blocks, producing citations at both the page and block levels. To implement this, we adapt and migrate the method proposed by \citet{DBLP:journals/corr/abs-2412-14457} to the multimodal RAG context to enable fine-grained block-level citation.

Although evaluation metrics for retrieval and generation are well-established, visual citation, as a novel application within RAG, still lacks dedicated evaluation methodologies. To address this gap, we propose an \textbf{automatic evaluation method for visual citation}. Specifically, we define the evaluation metrics, precision and recall, at both the page-level and block-level, and introduce two evaluation strategies: box-bounding and image-cropping.

We conduct extensive experiments and evaluations on FinRAGBench-V. For retrieval, we conduct experiments using four textual retrievers, such as Jina-ColBERT-V2 \cite{DBLP:journals/corr/abs-2408-16672},  and five Multilingual-E5-large \cite{DBLP:journals/corr/abs-2402-05672}; and multimodal ones, such as ColQwen2 \cite{DBLP:journals/corr/abs-2407-01449}, GME-Qwen2-VL-2B \cite {DBLP:journals/corr/abs-2412-16855}, and DSE-QWen2-2b-MRL-V1 \cite{DBLP:conf/emnlp/MaL0CL24}. For generation and citation, we employ seven proprietary Multimodal Large Language Models (MLLMs), such as GPT-4o \cite{DBLP:journals/corr/abs-2303-08774}, GPT-4V, and Gemini-2.0-Flash \cite{DBLP:journals/corr/abs-2507-06261}, and six open-source ones, such as Qwen2.5-VL-72B-Instruct \citet{Qwen2VL} and MiniCPM-o-2.6 \cite{yao2024minicpm}. 

Through the experiments, we derive several meaningful observations: (1) Multimodal retrievers outperform text-only ones by preserving information from charts and tables, avoiding information loss. (2) Current MLLMs handle text inference well but struggle with numerical reasoning on charts, tables, and multi-page inferences. (3) Multimodal RAG systems excel at page-level citation but struggle with block-level citation, highlighting challenges in precise attribution.
\begin{table*}[ht]
\centering
\footnotesize
\resizebox{\textwidth}{!}{
\setlength{\tabcolsep}{4pt}
\begin{tabular}{ccccccc}
\toprule
\textbf{Benchmark} & \textbf{Domain} & \textbf{RAG Corpus} & \textbf{Multimodal} & \textbf{Multi-Task} & \textbf{Multi-Page} & \textbf{Citation} \\
\midrule
\textbf{FinQA \cite{DBLP:conf/emnlp/ChenCSSBLMBHRW21}} & Finance & \xmark & \xmark & \xmark & \xmark & \xmark \\
\textbf{OmniEval \cite{DBLP:journals/corr/abs-2412-13018}} & Finance & \cmark & \xmark & \cmark & \xmark & \xmark \\
\textbf{EvoChart \cite{DBLP:conf/aaai/HuangLZW0ZL25}} & General & \xmark & \xmark & \xmark & \xmark & \xmark \\
\textbf{M3DocVQA \cite{DBLP:journals/corr/abs-2411-04952}} & General & \cmark & \cmark & \xmark & \cmark & \xmark \\
\textbf{VisDoMBench \cite{DBLP:journals/corr/abs-2412-10704}} & General & \cmark & \cmark & \xmark & \xmark & \xmark \\
\textbf{MME-Finance \cite{DBLP:journals/corr/abs-2411-03314}} & Finance & \xmark & \cmark & \cmark & \xmark & \xmark \\
\midrule
\textbf{FinRAGBench-V (Ours)} & Finance & \cmark & \cmark & \cmark & \cmark & \cmark \\
\bottomrule
\end{tabular}
}
\caption{Comparison of our benchmark with existing benchmarks.}
\label{tab:financial-benchmarks}
\end{table*}

In summary, our contributions are as follows:
\begin{itemize}
\item We construct FinRAGBench-V, a benchmark for visual RAG in the financial domain, featuring diverse real-world data sources for retrieval, a wide range of question types for generation, and visual citation for attribution.
\item We propose RGenCite, a comprehensive multimodal RAG baseline that combines retrieval, generation, and fine-grained visual citation. The model is required not only to generate answers from retrieved content, but also to provide page- and block-level visual citations as supporting evidence.
\item We propose an automatic evaluation method for visual citation. The method incorporates precision and recall metrics for citations at different levels, with evaluation approaches including box-bounding and image-cropping.
\item Extensive experiments reveal retriever differences, task-dependent model performance, and challenges in visual citation, validating FinRAGBench-V’s value for evaluating multimodal RAG in finance.
\end{itemize}

\section{Related Work}

\paragraph{Benchmarking Multimodal RAG.} 
Retrieval-Augmented Generation (RAG) has gained significant attention as an effective method of leveraging retrieval mechanisms to provide external knowledge to LLMs' generation \citep{DBLP:journals/corr/abs-2312-10997,DBLP:conf/nips/LewisPPPKGKLYR020,DBLP:journals/corr/abs-2308-07922, DBLP:conf/aaai/0011LH024,DBLP:journals/corr/abs-2407-11005,DBLP:conf/naacl/Saad-FalconKPZ24}. 
In the financial domain, where charts and graphs are essential, text-only RAG benchmarks often overlook critical information \citep{DBLP:conf/emnlp/ChenCSSBLMBHRW21,DBLP:journals/corr/abs-2412-13018}, highlighting the need for a multimodal RAG benchmark. Recent efforts on financial multimodal benchmarks exhibit several limitations, as summarized in Table~\ref{tab:financial-benchmarks}. EvoChart \citep{DBLP:conf/aaai/HuangLZW0ZL25} focuses solely on chart-based questions, lacking integration with textual and tabular information. \citet{DBLP:journals/corr/abs-2411-04952} and \citet{DBLP:journals/corr/abs-2412-10704} utilize real-world PDFs but support only limited question types. MME-Finance \citep{DBLP:journals/corr/abs-2411-03314} provides diverse financial questions, yet its reliance on isolated chart screenshots hinders document-level retrieval and fails to reflect the complexity of financial data.

\paragraph{Citation and Its Evaluation.}  
Citations play a crucial role in enhancing the credibility and interpretability of RAG systems \citep{DBLP:conf/acl/SlobodkinHCSD24,DBLP:journals/corr/abs-2311-03731,DBLP:conf/acl/Li0PMS24,DBLP:conf/emnlp/GaoYYC23}. While prior works focus on textual citations, \citet{DBLP:journals/corr/abs-2412-14457} introduce a coordinate-based method for multimodal citations. In specialized domains such as finance, where precise domain knowledge is essential, citation is particularly critical for RAG. Thus, we adapt this visual citation approach to the financial multimodal RAG setting and propose an automatic evaluation method for visual citation.
\section{Task Definition}
\begin{figure*}[t]
  \centering
  \includegraphics[clip=true,width=1\textwidth]{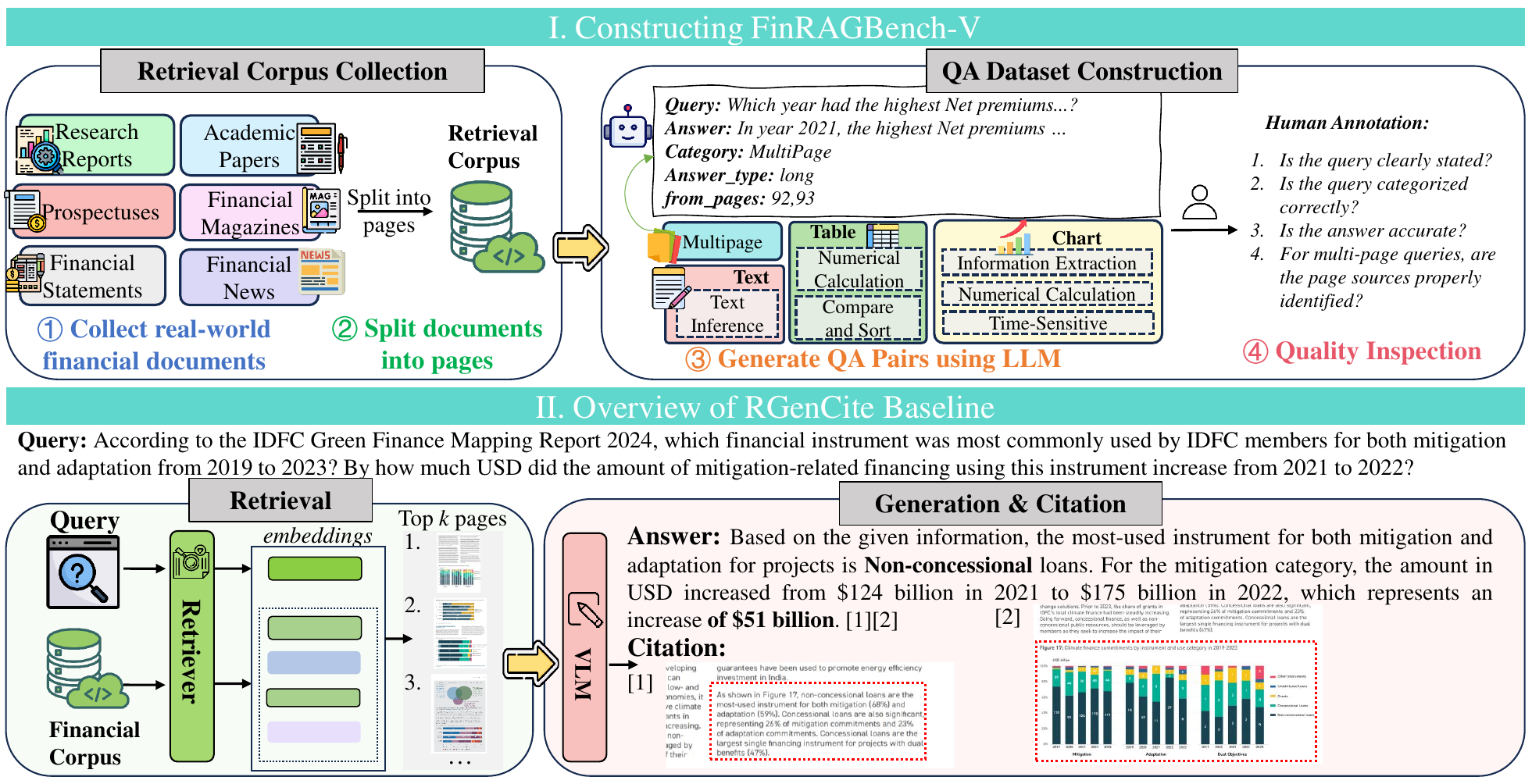}
  \caption{\textbf{I. Workflow of constructing FinRAGBench-V, including a retrieval corpus and a QA dataset}: \ding{172} collect real-world financial documents; \ding{173} split documents into pages; \ding{174} generate data using LLM;  \ding{175} quality inspection. \textbf{II. Overview of RGenCite Baseline}: including the retrieval stage, and generation-citation stage.}
  \label{fig:Task_workflow}
\end{figure*}
Our task contains two main phases: the construction of FinRAGBench-V, and the implementation of the RGenCite baseline, as shown in Figure \ref{fig:Task_workflow}.

In the first phase, given the raw documents collected from diverse sources, we first generate a retrieval corpus of pages, defined as $\mathcal{S}=\{p_1, p_2, ..., p_i, ...\}$, where $p_i$ represents the $i$th page. Based on the corpus, we generate the QA dataset, defined as $\mathcal{D}=\{d_1, d_2, ..., d_i, ...\}$, where each $d_i=(q_i, a_i, t_i, P_i)$, with $q_i$ being the question, $a_i$ the ground truth answer, $t_i$ the question type, and $P_i$ the set of corresponding page(s). So far, we have constructed the retrieval corpus and QA dataset.

The second phase comprises both the retrieval stage and the generation with citation stage. Given a question $q$, a retriever $R$ retrieves the top-k relevant pages $\{p_1, p_2, ..., p_k\}$ from the corpus $\mathcal{S}$. These pages, along with the question are then fed into a generator model $M$, which produces an answer $a$ accompanied by a set of citations $C=\{c_1, c_2, ..., c_i\}$. Each citation $c_i = (p_i, B_i)$ consists of a cited page $p_i$ and its corresponding supporting blocks $B_i = \{b_{i1}, b_{i2}, ..., b_{ij}\}$.

\section{The Construction of FinRAGBench-V}
As shown at the top of Figure \ref{fig:Task_workflow}, FinRAGBench-V consists of two components: a retrieval corpus and a QA dataset. This section outlines the construction process and provides detailed statistics.

\subsection{Retrieval Corpus Collection}
To build the retrieval corpus, we collect data from a variety of real-world financial document sources in both Chinese and English, as detailed in Appendix \ref{sec:sources}, including:

(1) \textbf{Research reports} collected from websites like Qianzhan.com, which provide in-depth financial analyses, for example the analysis of price trends over time using line charts; 

(2) \textbf{Financial statements of companies and banks} collected from the FinGLM \footnote{\url{https://tianchi.aliyun.com/competition/entrance/532164/introduction}}dataset and official company and bank websites, which provide annual financial data in tabular form;

(3) \textbf{Prospectuses} sourced from the BSCF \footnote{\url{https://www.modelscope.cn/datasets/BJQW14B/bs_challenge_financial_14b_dataset/}} dataset, providing information on companies going public, including financial data and business strategies, with rich tabular information;

(4) \textbf{Academic papers} offering theoretical and empirical insights into financial markets, economic models, and financial technologies, sourced from Journal of Financial and CNKI; 

(5) \textbf{Financial magazines} including respected outlets like the Financial Times, which offer reliable news, expert opinions, and financial analyses; 

(6) \textbf{Financial news} from websites like China Daily and Eastmoney.

We finally select 1,104 Chinese and 1,105 English documents from the aforementioned data sources (details in Table \ref{tab:dataset_stats}). Each document page is converted into a single image, resulting in a retrieval corpus of 60,780 Chinese and 51,219 English pages. By incorporating these diverse data types, we ensure that the retrieval corpus is both broad and reliable, providing a solid foundation for generating accurate and informative QA pairs. 

\begin{table}[h]
  \centering
  \resizebox{\columnwidth}{!}{
  \begin{tabular}{lcccc}
    \toprule
    \textbf{Data Source} & \textbf{Content Type} & \textbf{\#Docs} & \textbf{\#Pages} & \textbf{\#Avg. Pages} \\
    \midrule
    Research Reports          & Chart, Table, Text         & 219  & 8,583  & 52  \\
    Financial Statements      & Table, Text               & 408  & 38,004 & 376 \\
    Prospectuses             & Table, Text               & 41   & 539    & 13  \\
    Academic Papers          & Chart, Table, Text        & 311  & 1,912  & 10  \\
    Financial Magazines      & Chart, Text               & 191   & 9,958  & 131 \\
    Financial News           & Chart, Table, Text        & 1,039  & 1,784  & 3   \\
    \bottomrule
  \end{tabular}
  }
  \caption{Statistics of the corpus showing the types of document content, total document number, total pages, and average pages per document for each data source.}
  \label{tab:dataset_stats}
\end{table}

\subsection{QA Dataset Construction}
To construct the QA dataset, we follow a two-step process: first, we use a generator LLM to synthesize the QA pairs, and then conduct human annotation to ensure data quality. 
\subsubsection{QA Pairs Synthesis}
From the retrieval corpus, we select high-quality document pages and then generate a dataset using GPT-4o based on these pages, with predefined categories and carefully designed examples provided as prompts (provided in Appendix \ref{sec:prompts}). In terms of data scope, the dataset includes both single-page and multi-page questions; Regarding data format, it covers text, charts, and tables; As for answers, it contains both short and long ones; Considering the characteristics of financial domain, we further categorize the QA dataset into seven main categories as follows. Appendix \ref{sec:dataset} shows some examples.

\textbf{Text Inference}: This involves tasks like information extraction and summarization, such as deriving key insights or identifying specific details (e.g., financial data or trends) from text.

\textbf{Chart Information Extraction}: This involves extracting key metrics or features from charts, such as the percentage of a sector in a pie chart.

\textbf{Chart Numerical Calculations}: This involves performing numerical calculations based on charts, such as calculating the changes of interest rate.

\textbf{Chart Time-Sensitive Queries}: This involves handling time-based chart queries, such as identifying event timings, analyzing trends, and pinpointing data peaks and troughs, often focusing on how indicators evolve over time.

\textbf{Table Numerical Calculations}: Similar to chart calculations, this involves performing numerical operations on table data, such as calculating interest rate changes and summing costs, to derive insights.

\textbf{Table Comparison and Sorting}: This involves comparing and sorting table data, such as comparing financial indicators between entities, ranking them, or identifying the highest or lowest values.

\textbf{Multi-Page Queries}: This involves queries requiring information from multiple pages, such as extracting truncated tables or combining data from multiple charts to answer a single query.
\begin{figure}[t]
  \centering
  \includegraphics[clip=true,width=0.5\textwidth]{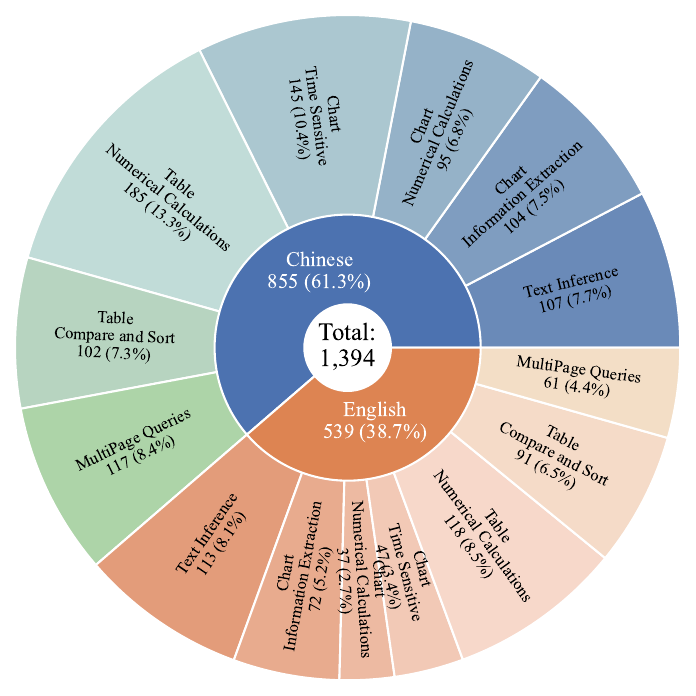}
  \caption{Statistics of Question Types in the Dataset.}
  \label{fig:The statistics and categories of the dataset}
\end{figure}

\subsubsection{Quality Inspection}
During the selection and annotation process, we adhere to several key principles to ensure the high quality and consistency of the dataset: examining the clarity of the questions and their correct categorization, verifying the accuracy of answers, and checking whether the page sources for multi-page queries are properly identified. The detailed annotation guideline is shown in Table \ref{tab:qa_annotation_guideline} of Appendix \ref{sec:guideline}. Based on these criteria, we carefully filter and refine the original 11,328 generated QA pairs, and ultimately obtaining a total of 1,394 pairs, consisting of 855 Chinese entries and 539 English entries. The statistics of each category are shown in Figure \ref{fig:The statistics and categories of the dataset}, the lengths statistics of the dataset are shown in Table \ref{tab:question_length_stats}.
\begin{table}[ht]
  \centering
  \resizebox{\columnwidth}{!}{
  \begin{tabular}{lcccc}
    \toprule
    \textbf{Category} & \textbf{Question} & \textbf{Answer (Overall)} & \textbf{Short Answer} & \textbf{Long Answer} \\
    \midrule
    \textbf{\#Avg. Length} & 12.23 & 10.17 & 5.37 & 47.56 \\
    \bottomrule
  \end{tabular}
  }
  \caption{Statistics of average token length of the dataset.}
  \label{tab:question_length_stats}
\end{table}
\section{RGenCite: Retrieval, Generation, and Visual Citation}
Based on our retrieval corpus and QA dataset, we develop the baseline system RGenCite, which covers both retrieval and generation, with visual citation seamlessly integrated into the generation stage, as illustrated at the bottom of Figure~\ref{fig:Task_workflow}.

\subsection{Retrieval}
During the retrieval stage, given a query $q$, the retrievers aim to identify the top-k relevant pages $\{p_1, p_2, ..., p_k\}$ from the corpus $S$. We explore various multimodal and textual retrievers and conduct a comprehensive evaluation of these two retrieval paradigms using multiple metrics.
\begin{figure*}[htbp]
  \centering
  \includegraphics[clip=true,width=1\textwidth]{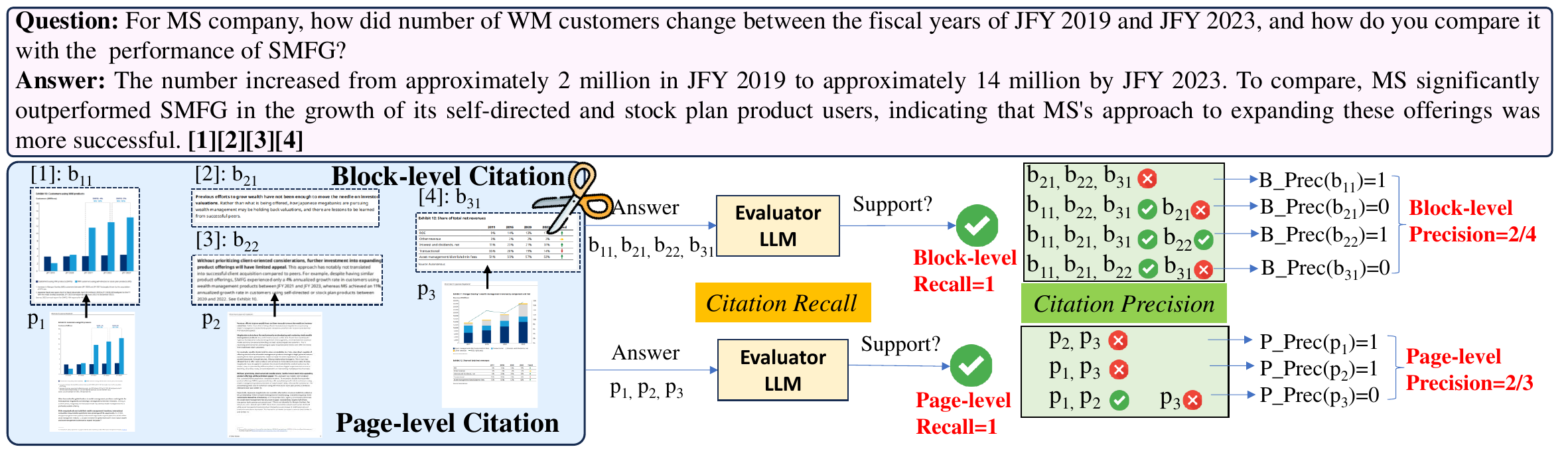}
  \caption{An example of the automatic evaluation of visual citation.}
  \label{fig:citation}
\end{figure*}
\subsection{Generation with Visual Citation}
During the generation stage, based on the retrieval result, the generator model $M$ is tasked with producing textual answer $a$ accompanied by visual citations $C$, given the query $q$. To enable the simultaneous generation of both answers and citations, we follow the visual citation method used in VISA \cite{DBLP:journals/corr/abs-2412-14457}. Specifically, we input both the question $q$ and the top-k relevant pages $\{p_1, p_2, ..., p_k\}$ into the generator $M$, instructing it to generate the answer $a$ while simultaneously producing both page-level and block-level citations. Each citation is denoted as $c_i=(p_i, \{b_{i1}, b_{i2},..., b_{ij}, ...\})$, where the page-level citation $p_i$ refers to the reference page, $\{b_{i1}, b_{i2},..., b_{ij},...\}$ represents the block-level citations, indicating the specific regions of the answer within the page. Each block-level citation $b_{ij}$ is represented as a set of coordinates, i.e., $b_{ij} = \left[x_1, y_1, x_2, y_2\right]$, where $(x_1,y_1)$ and $(x_2,y_2)$ denote the coordinates of the top-left corner and bottom-right corner of $b_{ij}$, respectively. Detailed output format is in Appendix \ref{sec:prompts}. 

\section{Evaluation Metrics}
After implementation, we evaluate the RGenCite baseline from three perspectives: retrieval, generation, and visual citation, with citation quality assessed using our proposed evaluation method.
\subsection{Retrieval Quality}
To evaluate the performance of both multimodal and textual retrievers, we adopt several evaluation metrics, namely nDCG@k (for k = 5, 10), Recall@k (for k = 5, 10), and MRR@k (k = 10), which respectively capture ranking quality, retrieval coverage, and early relevance.
\subsection{Answer Accuracy}
To evaluate MLLMs' ability to generate accurate responses based on visual elements, we use the rule-based metric ROUGE. Additionally, we employ GPT-4o to assess the metric Acc, determining whether the generated responses align with the ground truths and are consistent with the visual context. The evaluation prompt is in Appendix \ref{sec:prompts}.
\subsection{Citation Quality}
To evaluate the visual citation quality of MLLMs, we introduce two automatic evaluation metrics: recall and precision. These metrics are applied at both the page-level and the block-level, using two distinct citation evaluation approaches: box-bounding and image-cropping. The effectiveness of our automatic citation evaluation methods is demonstrated in Section \ref{section:consistency}.

\paragraph{Citation Metrics.} Inspired by \citet{DBLP:conf/emnlp/GaoYYC23}, we evaluate both page-level and block-level citations using the following two metrics:

\textbf{Recall} evaluates whether the cited images are sufficient to support the answer. If the union of the citation set $C=\{c_1,c_2,...,c_n\}$ of an answer $a$ sufficiently support $a$, the recall is assigned $1$; otherwise, it is assigned $0$, defined in Equation \ref{eq:recall}:

\begin{equation}
\text{recall}(C, a) = \begin{cases} 
1 & \text{if } \bigcup_{c_i \in C} c_i \text{ supports } a, \\
0 & \text{otherwise}.
\end{cases}
\label{eq:recall}
\end{equation}

\definecolor{mmColor}{RGB}{220,240,255}  
\definecolor{textColor}{RGB}{255,235,230} 
\begin{table*}
  \centering
  \resizebox{\textwidth}{!}{
  \begin{tabular}{lccccc|ccccc}
  \toprule
   \multirow{2}{*}{\textbf{Retriever}} & \multicolumn{5}{c}{\textbf{Chinese}} & \multicolumn{5}{c}{\textbf{English}} \\
    \cline{2-6} \cline{7-11}
    & \textbf{nDCG@5} & \textbf{nDCG@10} & \textbf{Recall@5} & \textbf{Recall@10} & \textbf{MRR@10} 
    & \textbf{nDCG@5} & \textbf{nDCG@10} & \textbf{Recall@5} & \textbf{Recall@10} & \textbf{MRR@10} \\
    
    \rowcolor{mmColor} \multicolumn{11}{c}{\textbf{Multimodal Retrievers}} \\
    \rowcolor{mmColor!60} ColQwen2 & \textbf{78.53} & \textbf{79.76} & \textbf{86.46} & \textbf{90.13} & \textbf{77.80} 
                                 & \textbf{67.90} & \textbf{70.00} & \textbf{79.64} & \textbf{85.86} & \textbf{65.54} \\
    \rowcolor{mmColor!60} GME-Qwen2-VL-7B & 74.55 & 76.04 & 84.80 & 89.35 & 72.80 & 58.06 & 60.94 & 68.95 & 77.56 & 56.23 \\
    \rowcolor{mmColor!60} GME-Qwen2-VL-2B & 63.49 & 79.66 & 73.14 & 79.66 & 64.99 & 53.83 & 56.22 & 64.46 & 71.56 & 52.10 \\
    \rowcolor{mmColor!60} DSE-Qwen2-2b-MRL-V1 & 61.16 & 63.07 & 69.71 & 75.62 & 60.15 & 62.37 & 64.70 & 74.44 & 81.50 & 60.03 \\
    \rowcolor{mmColor!60} VisRAG-Ret & 55.17 & 57.81 & 66.40 & 74.47 & 53.60 & 51.56 & 54.99 & 64.93 & 75.40 & 49.48 \\
    
    \rowcolor{textColor} \multicolumn{11}{c}{\textbf{Text Retrievers}} \\
    \rowcolor{textColor!60} BGE-M3 & 31.49 & 33.09 & 37.92 & 42.71 & 29.93 & 23.90 & 25.87 & 31.17 & 36.36 & 22.21 \\
    \rowcolor{textColor!60} Multilingual-E5-large & 28.45 & 30.41 & 35.12 & 41.07 & 26.97 & 22.70 & 24.83 & 28.57 & 35.06 & 21.64 \\
    \rowcolor{textColor!60} Jina-ColBERT-V2 & 24.61 & 25.93 & 28.82 & 33.02 & 23.68 & 16.72 & 18.56 & 21.52 & 27.27 & 15.88 \\
    \rowcolor{textColor!60} BM25 & 11.39 & 12.65 & 14.70 & 18.67 & 10.79 & 18.26 & 21.63 & 26.35 & 31.54 & 18.52 \\
    
    \bottomrule
  \end{tabular}
  }
  \caption{Retrieval results for both Chinese and English in percentage. The best results are highlighted in \textbf{bold}.}
  \label{tab:retrieval_results}
\end{table*}

\begin{table*}
  \centering
  \resizebox{\textwidth}{!}{
  \begin{tabular}{lcccccc|cccccc}
  \toprule
    \multirow{2}{*}{\textbf{Model}} & \multicolumn{6}{c}{\textbf{Chinese}} & \multicolumn{6}{c}{\textbf{English}} \\
    \cmidrule(lr){2-7} \cmidrule(lr){8-13}
    & \textbf{ROUGE} & \textbf{Acc} & \textbf{P\_Rec} & \textbf{P\_Prec} & \textbf{B\_Rec} & \textbf{B\_Prec} & \textbf{ROUGE} & \textbf{Acc} & \textbf{P\_Rec} & \textbf{P\_Prec} & \textbf{B\_Rec} & \textbf{B\_Prec} \\
    \hline
    \rowcolor{grayColor}\multicolumn{13}{c}{\textbf{Proprietary MLLMs}} \\
    \hline
    \rowcolor{color1!20} o4-mini & \textbf{38.55} & \textbf{58.13} & 78.01 & 75.77 & 54.74 & 48.20 & \cellcolor{color1!30} \textbf{40.21} & \cellcolor{color1!30}\textbf{69.20} & \cellcolor{color1!30}75.32 & \cellcolor{color1!30}75.32 & \cellcolor{color1!30}\textbf{60.11} & \cellcolor{color1!30}\textbf{55.75} \\
    \rowcolor{color1!20} GPT-4o & 26.82 & 33.26 & 92.15 & 87.27 & \underline{61.01} & \underline{52.80} & \cellcolor{color1!30} \underline{24.66} & \cellcolor{color1!30}43.41 & \cellcolor{color1!30}\textbf{89.98} & \cellcolor{color1!30}\underline{81.81} & \cellcolor{color1!30}\underline{54.17} & \cellcolor{color1!30}\underline{44.66} \\
    \rowcolor{color1!20} GPT-4V & 26.38 & 31.70 & \textbf{93.10} & \underline{88.56} & \textbf{61.29} & \textbf{52.88} & \cellcolor{color1!30} 22.76 & \cellcolor{color1!30}44.71 & \cellcolor{color1!30}89.24 & \cellcolor{color1!30}80.54 & \cellcolor{color1!30}53.43 & \cellcolor{color1!30}42.69 \\
    \rowcolor{color1!20} GPT-4o-mini & 19.46 & 19.53 & 78.07 & 56.08 & 24.68 & 16.17 & \cellcolor{color1!30} 16.21 & \cellcolor{color1!30}28.94 & \cellcolor{color1!30}60.30 & \cellcolor{color1!30}41.20 & \cellcolor{color1!30}22.63 & \cellcolor{color1!30}13.23 \\
    \rowcolor{color2!20} Gemini-1.5-Flash & 18.18 & 21.34 & 69.58 & 67.10 & 20.62 & 16.80 & \cellcolor{color2!30} 16.24 & \cellcolor{color2!30}26.72 & \cellcolor{color2!30}72.17 & \cellcolor{color2!30}66.71 & \cellcolor{color2!30}25.97 & \cellcolor{color2!30}21.05 \\
    \rowcolor{color2!20} Gemini-2.0-Flash & \underline{28.00} & \underline{41.40} & \underline{92.87} & \textbf{89.58} & 34.07 & 29.29 & \cellcolor{color2!30} 21.83 & \cellcolor{color2!30}\underline{46.01} & \cellcolor{color2!30}\underline{89.61} & \cellcolor{color2!30}\textbf{85.22} & \cellcolor{color2!30}20.41 & \cellcolor{color2!30}17.23 \\
    \rowcolor{color3!20} Claude-3.5-Sonnet & 21.87 & 32.67 & 59.48 & 55.54 & 31.81 & 28.62 & \cellcolor{color3!30} 20.92 & \cellcolor{color3!30}43.41 & \cellcolor{color3!30}79.78 & \cellcolor{color3!30}77.99 & \cellcolor{color3!30}36.73 & \cellcolor{color3!30}34.49 \\
    \hline
    \rowcolor{grayColor} \multicolumn{13}{c}{\textbf{Open-Source MLLMs}} \\
    \hline
    \rowcolor{color4!20} Qwen2-VL-72B-Instruct & 22.83 & 30.41 & 58.25 & 51.31 & 10.64 & 9.49 & \cellcolor{color4!30} \textbf{25.85} & \cellcolor{color4!30}25.97 & \cellcolor{color4!30}53.80 & \cellcolor{color4!30}43.68 & \cellcolor{color4!30}7.42 & \cellcolor{color4!30}5.91 \\
    \rowcolor{color4!20} Qwen2.5-VL-7B-Instruct & 22.19 & 30.06 & \underline{65.38} & \underline{62.27} & 9.71 & 8.19 & \cellcolor{color4!30} 19.47 & \cellcolor{color4!30}\underline{36.36} & \cellcolor{color4!30}51.21 & \cellcolor{color4!30}\underline{49.25} & \cellcolor{color4!30}18.74 & \cellcolor{color4!30}15.72 \\
    \rowcolor{color4!20} Qwen2.5-VL-32B-Instruct & \textbf{25.89} & \underline{34.66} & \textbf{74.71} & \textbf{65.95} & \textbf{33.37} & \underline{23.45} & \cellcolor{color4!30} 21.33 & \cellcolor{color4!30}30.05 & \cellcolor{color4!30}\underline{59.00} & \cellcolor{color4!30}48.03 & \cellcolor{color4!30}\underline{35.44} & \cellcolor{color4!30}\underline{24.47} \\
    \rowcolor{color4!20} Qwen2.5-VL-72B-Instruct & \underline{25.12} & \textbf{36.02} & 61.17 & 55.72 & \underline{32.75} & \textbf{28.54} & \cellcolor{color4!30} \underline{21.98} & \cellcolor{color4!30}\textbf{38.03} & \cellcolor{color4!30}\textbf{68.09} & \cellcolor{color4!30}\textbf{63.93} & \cellcolor{color4!30}\textbf{39.52} & \cellcolor{color4!30}\textbf{35.03} \\
    \rowcolor{color5!20} MiniCPM-o-2.6 & 13.15 & 11.58 & 60.94 & 57.68 & 2.81 & 2.48 & \cellcolor{color5!30} 18.32 & \cellcolor{color5!30}9.83 & \cellcolor{color5!30}37.29 & \cellcolor{color5!30}36.30 & \cellcolor{color5!30}0.74 & \cellcolor{color5!30}0.46 \\
    \rowcolor{color6!20} Phi-3.5-V-Instruct & 5.14 & 4.55 & 35.91 & 34.19 & 3.39 & 2.72 & \cellcolor{color6!30} 6.70 & \cellcolor{color6!30}6.86 & \cellcolor{color6!30}24.12 & \cellcolor{color6!30}22.35 & \cellcolor{color6!30}0.74 & \cellcolor{color6!30}0.58 \\
    \bottomrule
  \end{tabular}
  }
  \caption{Results for generation and citation on FinRAGBench-V in percentage. For both proprietary models and open-source models, the best result is shown in \textbf{bold}, and the second-best is \underline{underlined}.}
  \label{tab:Results for Generation and Citation}
\end{table*}
\textbf{Precision} evaluates the proportion of citations in the citation set $C$ that are essential for supporting an answer. Specifically, the citation $c_i$ is considered irrelevant if and only if $c_i$ cannot independently support the answer, and the union of all other citations $\{c_1, c_2, ..., c_{i-1}, c_{i+1},...\}$ in $C$ is sufficient to support the answer $a$, as described in Equation \ref{eq:irrel}:
\begin{equation}
\text{irrel}(C, c_i, a) = \left( c_i \nrightarrow a \right) \land(\left(C \setminus \{c_i\}) \rightarrow a \right) 
\label{eq:irrel}
\end{equation}
Thus, the citation precision of the citation set $C$ for answer $a$ is defined as the proportion of non-irrelevant citations in $C$, as shown in Equation \ref{eq:prec}:
\begin{equation}
\text{precision}(C, a) = \frac{|C \setminus \{c_i \mid \text{irrel}(C, c_i, a) = 1\}|}{|C|}
\label{eq:prec}
\end{equation}
It should be noted that the precision of each citation is evaluated only when the recall of the citation set it belongs to is $1$; otherwise, i is set to $0$. 

\paragraph{Citation Evaluation.}
The citation quality is evaluated using the aforementioned metrics at two different levels: page-level and block-level, as shown in Figure \ref{fig:citation}, denoted as: \textit{P\_Rec}, \textit{P\_Prec}, \textit{B\_Rec}, and \textit{B\_Prec}. Moreover, we use two evaluation approaches: box-bounding and image-cropping, to assess the citation quality. As shown in Appendix \ref{sec:box_crop}, the former draws bounding boxes around relevant regions based on the citation coordinates, while the latter directly crops the cited image blocks accordingly. In both cases, we introduce an evaluator MLLM to determine citation quality. Through experiments in Section~\ref{section:consistency}, we find that image-cropping yields higher alignment with Intersection over Union (IoU) scores and human judgments, and therefore it is used as the default approach in subsequent evaluations.

\section{Experiments and Results}
We evaluate both the retrieval stage and the generation stage with citation using the aforementioned metrics. For retrieval, we assess both multimodal and textual retrievers. For generation, we use the best retriever to provide the top-$k$ pages ($k=10$) as input, comparing the performance of proprietary and open-source MLLMs across different tasks.

\subsection{Basic Settings}
\begin{table*}[ht]
\centering
\setlength{\tabcolsep}{4pt} 
\resizebox{0.9\textwidth}{!}{
\footnotesize
\begin{tabular}{cccccccc}
\toprule
\multirow{2}{*}{\textbf{Eval Approach}} & \multirow{2}{*}{\textbf{Eval Model}} 
& \multicolumn{3}{c}{\textbf{Consistency with IoU}} 
& \multicolumn{3}{c}{\textbf{Consistency with Human Eval}} \\
\cmidrule(lr){3-5} \cmidrule(lr){6-8}
 & & \textbf{Pearson} & \textbf{Spearman} & \textbf{Kendall}
   & \textbf{Pearson} & \textbf{Spearman} & \textbf{Kendall} \\
\midrule
\multirow{6}{*}{image-cropping}
 & GPT-4o & \textbf{65.06} & \textbf{63.08} & \textbf{54.58} &\textbf{68.01}  &\textbf{64.03}  &\textbf{57.37}  \\
 & GPT-4v & 63.27 & 61.49 & 53.21 &64.78  &60.98  &54.50  \\
 & GPT-4-turbo & 52.44 & 54.66 & 46.87 & 57.56 &54.82  &48.70  \\
 & Gemini-1.5-Flash & 53.55 & 50.47 & 43.59 &50.39  &47.01  &41.99  \\
 & Gemini-2.0-Flash & 54.18 & 53.89 & 46.17 &60.09  &57.86  & 51.42 \\
\midrule
box-bounding & GPT-4o & 7.28 & 9.19 & 8.14 & 12.30 &12.80  &11.29  \\
\bottomrule
\end{tabular}
}
\caption{Consistency of automatic citation evaluation methods with IoU and human evaluation in percentages.}
\label{tab:consistency}
\end{table*}

\paragraph{Retrieval.}
During the retrieval phase, we explore both multimodal retrievers alongside textual ones. (1) \textbf{Multimodal retrievers}: We evaluate five models, namely ColQwen2 \cite {DBLP:journals/corr/abs-2407-01449}, GME-Qwen2-VL-2B \cite {DBLP:journals/corr/abs-2412-16855}, GME-Qwen2-VL-7B, DSE-QWen2-2b-MRL-V1 \cite{DBLP:conf/emnlp/MaL0CL24}, and VisRAG-Ret \cite{DBLP:journals/corr/abs-2410-10594}, to assess their effectiveness in retrieving relevant content from multimodal pages. (2) \textbf{Text retrievers}: We use Marker \cite{paruchuri_marker} for OCR-based text extraction. Subsequently, we test four text retrievers, namely BM25, Jina-ColBERT-V2 \cite{DBLP:journals/corr/abs-2408-16672}, BGE-M3 \cite{DBLP:journals/corr/abs-2402-03216}, and Multilingual-E5-large \cite{DBLP:journals/corr/abs-2402-05672}, evaluating their effectiveness in retrieving relevant information from the extracted texts.

\paragraph{Generation with Visual Citation}
In the generation phase, we conduct experiments on both proprietary and open-source MLLMs. The former consists of o4-mini, GPT-4o \cite{DBLP:journals/corr/abs-2303-08774}, GPT-4V, GPT-4o-mini, Gemini-1.5-Flash \cite{DBLP:journals/corr/abs-2403-05530}, Gemini-2.0-Flash \cite{DBLP:journals/corr/abs-2507-06261}, and Claude-3.5-Sonnet-20240620 \cite{anthropic2024claude35sonnet}; while the later comprises Qwen2-VL-72B-Instruct \cite{Qwen2VL}, Qwen2.5-VL-7B-Instruct, Qwen2.5-VL-32B-Instruct, Qwen2.5-VL-72B-Instruct, Phi-3.5-vision-instruct \cite{DBLP:journals/corr/abs-2404-14219}, and MiniCPM-o-2.6 \cite{yao2024minicpm}. The prompt for generation is in Appendix \ref{sec:prompts}, more details are in Appendix \ref{sec:resource_usage}.
\subsection{Main Results}
\paragraph{Retrieval.} In the retrieval stage, we observe that \textbf{multimodal retrievers significantly outperform textual retrievers across all metrics}. As shown in Table~\ref{tab:retrieval_results}, ColQwen2 achieves a recall@10 of 90.13 (Chinese) and 85.86 (English), whereas the best textual retriever, BGE-M3, reaches only 42.71 and 36.36, respectively. This highlights the effectiveness of multimodal retrievers in handling complex financial data involving charts and tables.

\begin{figure}[h]
  \centering
  \includegraphics[clip=true,width=\columnwidth]{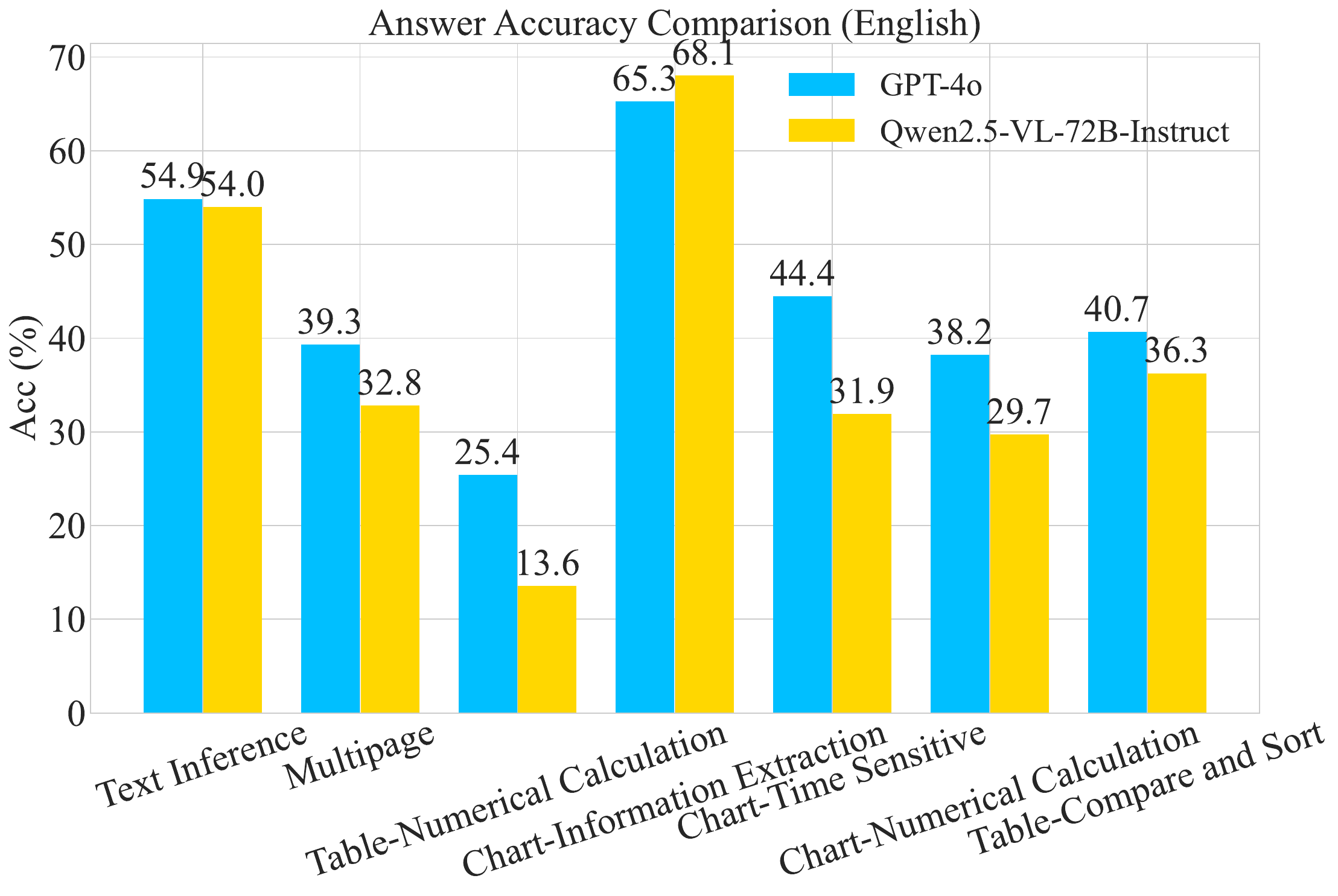}
  \caption{The comparison of answer accuracy between different question categories.}
  \label{fig:category_compare}
\end{figure}

\paragraph{Generation.} From Table~\ref{tab:Results for Generation and Citation}, we observe the following findings: (1) \textbf{Proprietary LLMs outperform their open-source counterparts}, underscoring the challenges that open-source MLLMs face in handling complex multimodal tasks. (2) \textbf{Different MLLMs show varying strengths on Chinese and English datasets.} Concretely, models such as GPT-4o, GPT-4V, Gemini-2.0-Flash, and Claude-3.5-Sonnet perform significantly better on English data, whereas Qwen2.5-VL-72B-Instruct and Qwen2-VL-72B-Instruct demonstrate balanced and even superior performance on Chinese data. (3) Task-wise analysis on FinRAGBench-V (Figure~\ref{fig:category_compare}) shows that \textbf{MLLMs excel at text inference and direct information extraction, but still struggle with numerical calculations and multi-page inference}. These observations suggest that complex visual reasoning tasks in specialized domains like finance remain a key challenge for current MLLMs. Some case studies on the typical errors are shown in Appendix \ref{section:case study}.
\paragraph{Visual Citation.} In terms of citation, Table \ref{tab:Results for Generation and Citation} shows that \textbf{most MLLMs perform well in page-level citations}, demonstrating their ability to accurately identify relevant pages from the provided references. However, \textbf{block-level citation remains difficult}, especially for open-source MLLMs. This highlights the challenge of attributing information to specific regions within a page, and suggests that many open-source MLLMs still struggle with precise citation generation. It also underscores the ongoing challenge of achieving accurate visual attribution within images, especially when pinpointing specific content blocks.

\subsection{Consistency between Automatic Citation Evaluation with Human Evaluations}\label{section:consistency}
To validate our automatic citation evaluation method, we measure its alignment with the following two human evaluation methods.

\paragraph{IoU-based Human Evaluation.} We employ the \textit{labelImg\footnote{https://github.com/HumanSignal/labelImg}} tool to manually annotate citation regions, which serve as the visual ground truth. The Intersection over Union (IoU) between predicted and annotated boxes is computed to quantify geometric overlap. Although intuitive, this metric has notable limitations for evaluating citation grounding quality, as it can be influenced by factors such as blank space within bounding boxes or missing key information that still yields a high IoU score.

\paragraph{Rating-based Human Evaluation.} To complement IoU, we use human ratings of the predicted citations on a 0–5 scale, considering factors such as page and block relevance, offset from ground truth, and the inclusion of redundant or irrelevant content. This provides a more nuanced and semantically meaningful assessment of citation quality. The guideline for rating is shown in Table \ref{tab:citation_annotation_guideline} of Appendix \ref{sec:guideline}.

As shown in Table \ref{tab:consistency}, we evaluate the citation performance of Qwen2.5-VL-72B using our automatic citation method across multiple variants, and assess its consistency with IoU scores and human ratings via Pearson, Spearman, and Kendall correlations coefficients. The image-cropping approach achieves Pearson correlations of 65.06 (with IoU) and 68.01 (with human ratings), demonstrating its effectiveness. In contrast, the box-bounding approach underperforms due to noise introduced by redundant visual content. Accordingly, we adopt GPT-4o with image-cropping in our experiments.

\section{Conclusion}
In this paper, we introduce FinRAGBench-V, a benchmark designed for multimodal RAG with visual citations in the financial domain, covering a retrieval corpus collected from diverse real-world financial documents and a QA dataset focusing on a wide range of financial tasks. Through extensive experiments, FinRAGBench-V exposes limitations of MLLMs and serves as a valuable resource to guide future improvements in visual RAG systems.

\section*{Limitations}
Despite the comprehensive experiments conducted in FinRAGBench-V that have provided valuable insights, our work still has limitations. Specifically, we did not train a dedicated model for multimodal RAG in the financial domain. Future work should address this by developing models tailored to the unique challenges of financial multimodal RAG, thereby enhancing the applicability and effectiveness of our benchmark.

\section*{Acknowledgement}
We thank the anonymous reviewers for their helpful comments on this paper. This work is supported by the National Natural Science Foundation of China (No. 62476010) and National Natural Science Foundation of China (No. 62272008). 
\bibliography{custom}

\appendix
\section{Prompts for QA Pairs Construction, Generation, and Evaluations}\label{sec:prompts}
We provide the prompts for constructing QA paris, generating answer with visual citations, and the evaluation on the answer and citations, shown in Table \ref{tab:Prompt for Constructing QA Dataset}, \ref{tab:Prompt for Generation and Citation}, \ref{tab:acc_prompt}, \ref{tab:citation_prompt_page}, \ref{tab:citation_prompt_bounding_boxes}, \ref{tab:citation_prompt_image_cropping}.

\section{Examples of Six Real-World Data Sources of Retrieval Corpus}\label{sec:sources}
\begin{figure}[p]
  \centering
  \includegraphics[clip=true,width=\columnwidth]{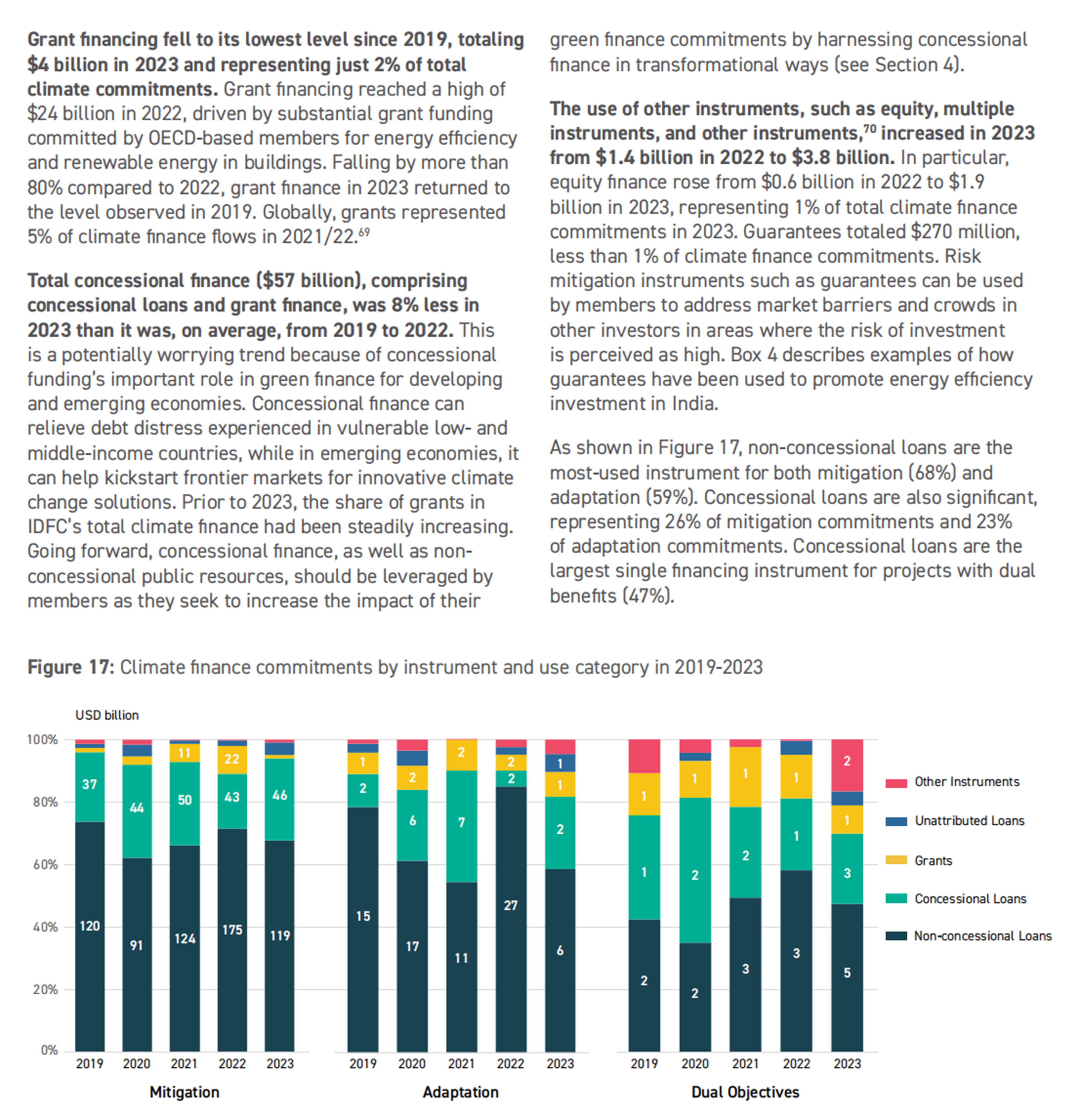}
  \caption{An example of research report}
  \label{fig:research_report}
\end{figure}

\begin{figure}[p]
  \centering
  \includegraphics[clip=true,width=\columnwidth]{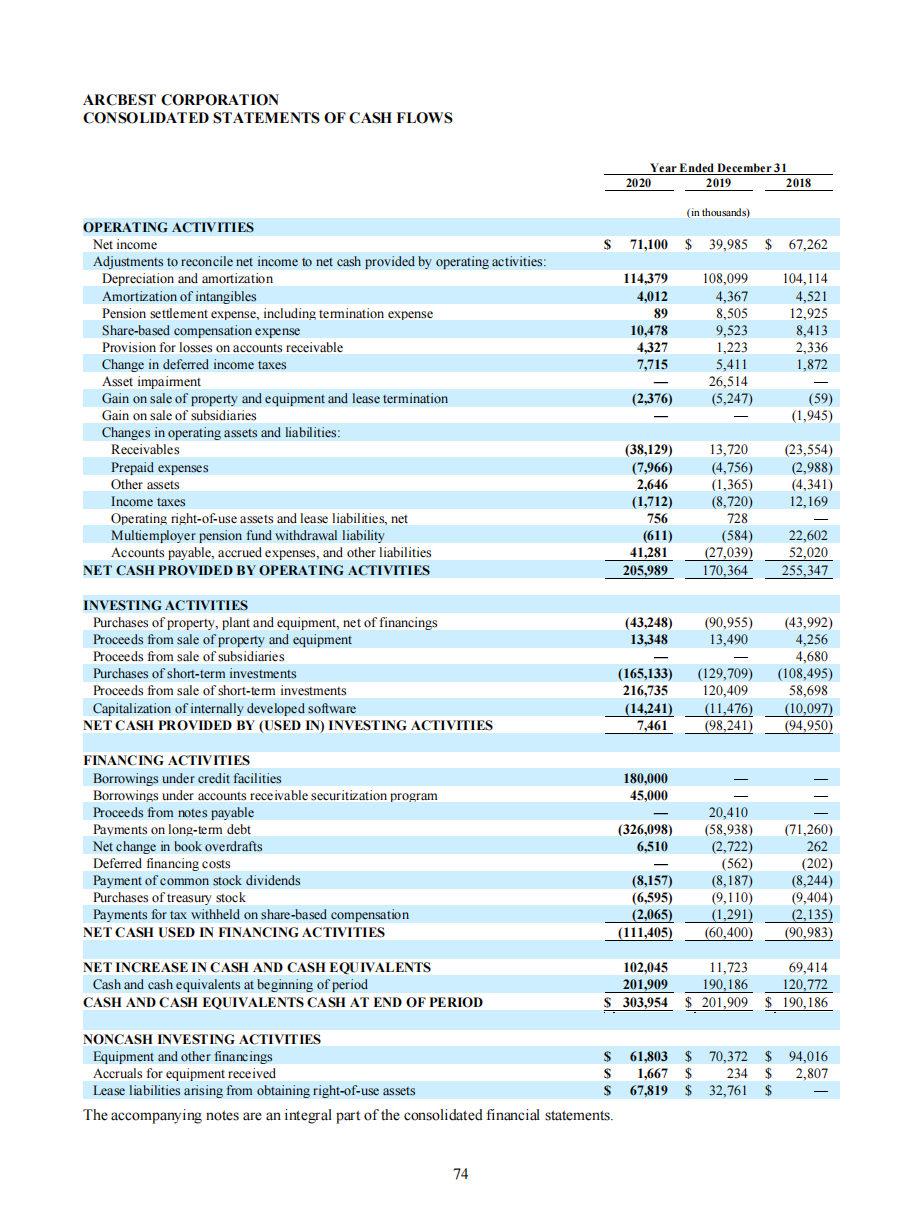}
  \caption{An example of financial statements}
  \label{fig:financial_statement}
\end{figure}

\begin{figure}[p]
  \centering
  \includegraphics[clip=true,width=\columnwidth]{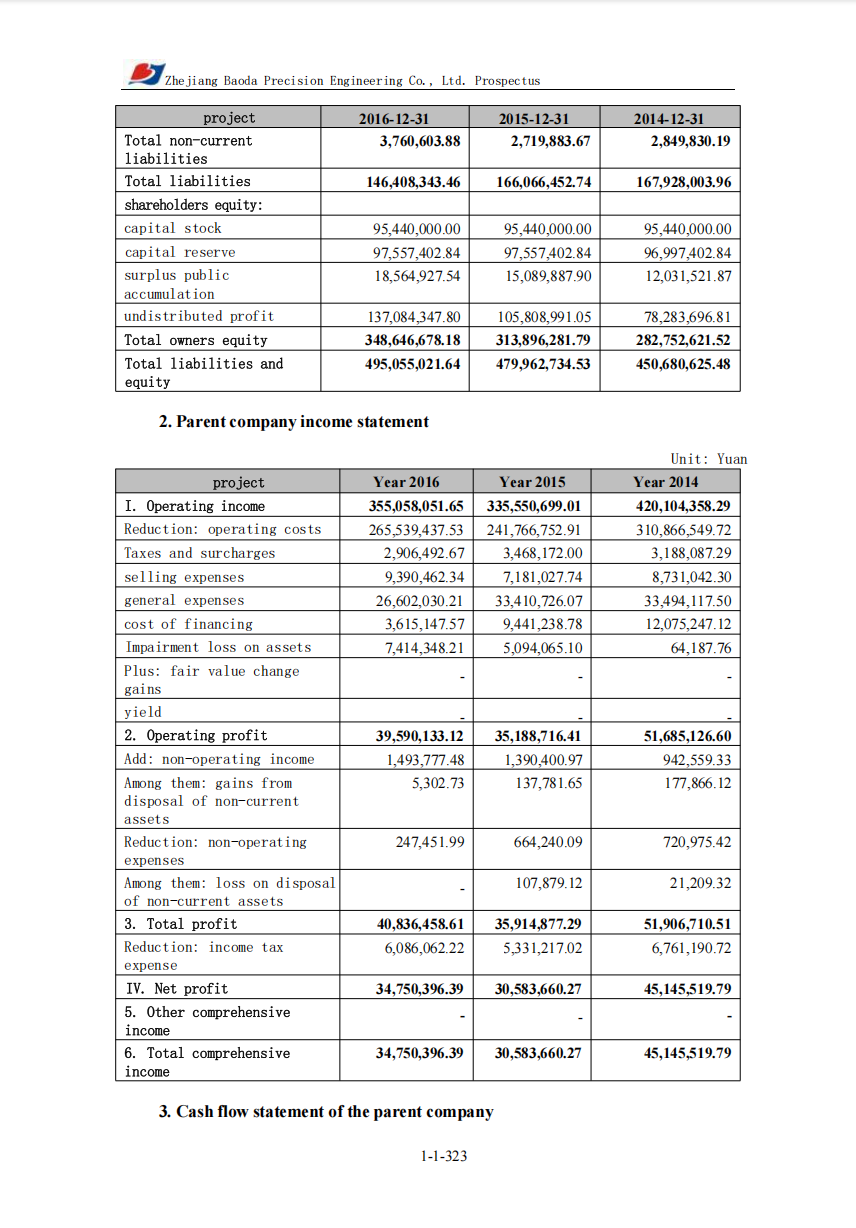}
  \caption{An example of prospectus}
  \label{fig:prospectus}
\end{figure}

\begin{figure}[p]
  \centering
  \includegraphics[clip=true,width=\columnwidth]{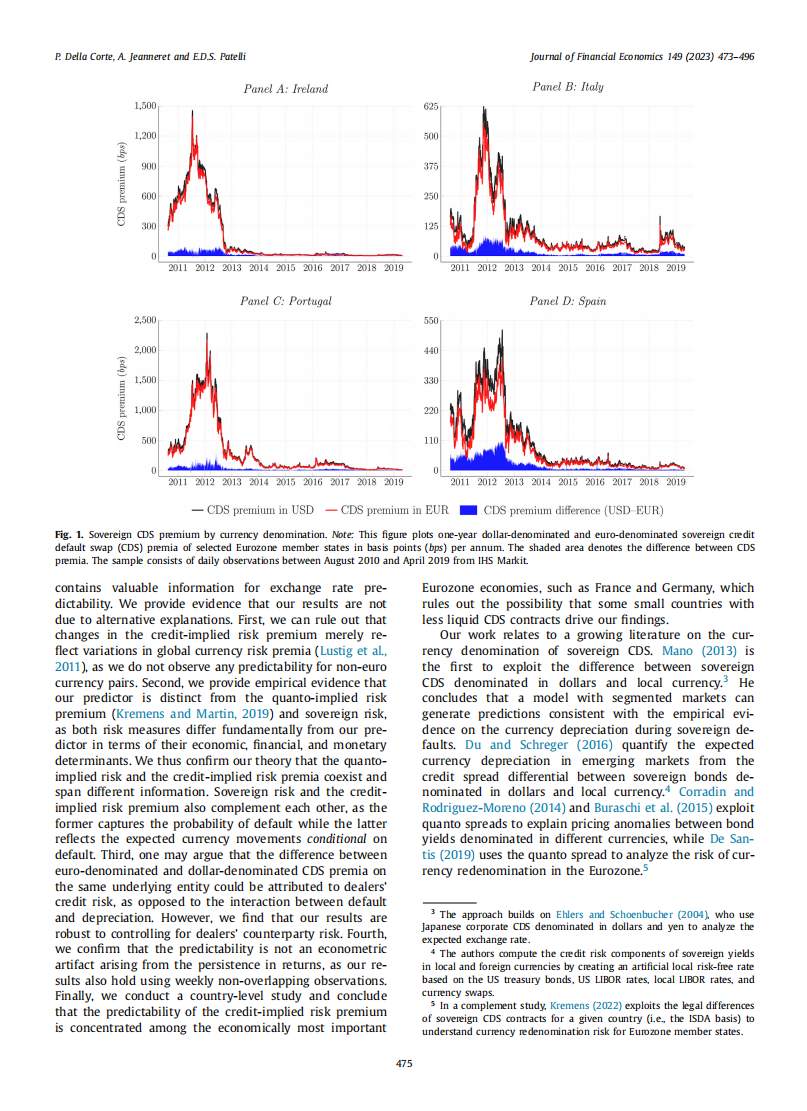}
  \caption{An example of finance-related academic paper}
  \label{fig:paper}
\end{figure}

\begin{figure}[p]
  \centering
  \includegraphics[clip=true,width=\columnwidth]{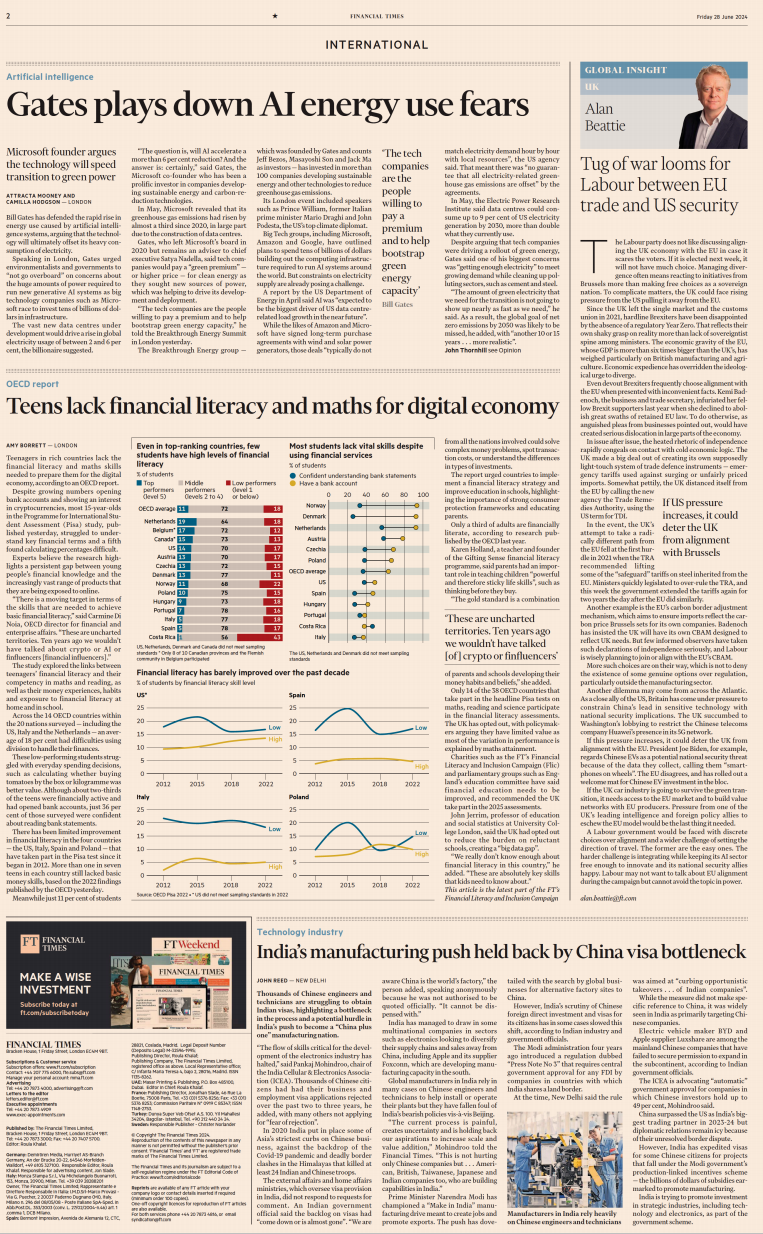}
  \caption{An example of financial magazine}
  \label{fig:magazine}
\end{figure}

\begin{figure}[p]
  \centering
  \includegraphics[clip=true,width=\columnwidth]{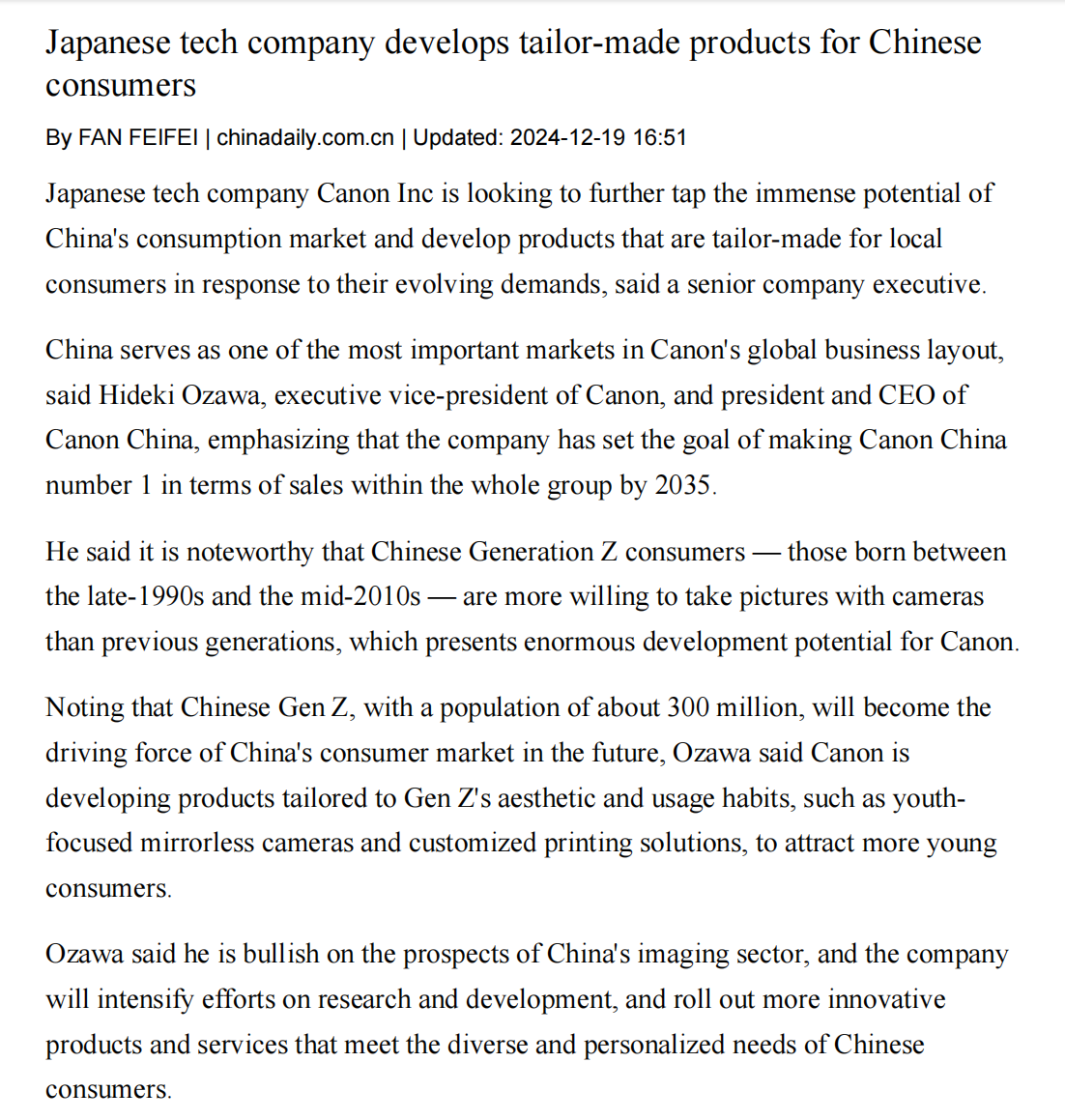}
  \caption{An example of financial news}
  \label{fig:news}
\end{figure}

In this section, we provide an example for each data source, illustrating the construction of our courpus, shown in Figure \ref{fig:research_report}, \ref{fig:financial_statement}, \ref{fig:prospectus}, \ref{fig:paper}, \ref{fig:magazine}, \ref{fig:news}.

\section{Examples of Seven Categories of QA Dataset}\label{sec:dataset}
In this section, we provide an example for each category of questions, shown in Table \ref{tab:text inference}, \ref{tab:text inference2} \ref{tab:chart info extract}, \ref{tab:chart info extract2}, \ref{tab:chart numerical}, \ref{tab:chart numerical2}, \ref{tab:chart time}, \ref{tab:chart time2}, \ref{tab:table-numerical}, \ref{tab:table-numerical2}, \ref{tab:table-compare}, \ref{tab:table-compare2}, \ref{tab:multipage}, \ref{tab:multipage2}.

\subsection{Text Inference:}
This category involves tasks such as summarization and information extraction from text. For example, deriving key insights from large volumes of text or identifying specific pieces of information, such as financial data or trends, within the content.

\subsection{Chart-Information Extraction}
This category focuses on extracting important metrics or features from charts. For example, it involves determining the exact percentage of a sector in a pie chart. 

\subsection{Chart-Numerical Calculations}
In this category, the focus is on performing numerical calculations based on the data presented in charts. Tasks include calculating the change of interest rates, summing up costs, and evaluating the percentage point increase in market share, among others.

\subsection{Chart-Time Sensitive}
This category addresses time-based queries related to charts. It includes identifying the timing of specific events, analyzing trends over time, pinpointing the peaks and troughs in the data, etc. These queries often involve examining how certain indicators evolve and identifying key moments in time.

\subsection{Table-Numerical Calculations}
Similar to chart calculations, this category involves performing numerical operations on the data presented in tables. Common tasks include calculating the change of interest rates, summing up costs, etc. These calculations help derive meaningful insights from tabular data. 

\subsection{Table-Comparison and Sorting}
This category focuses on comparing and sorting data within tables. It includes comparing financial indicators such as revenue or cost between different entities, as well as ranking them based on specific criteria. Tasks may also involve identifying the highest or lowest values among multiple entries.

\subsection{Multi-page Queries}
This category deals with queries that concern information from multiple pages. It includes tasks that span across text, tables, or charts split across pages. For example, it involves extracting truncated tables from different pages or interpreting information from multiple charts that need to be combined to answer a single query.

\section{Example for Visual Citation and the Two Evaluation Methods}\label{sec:box_crop}
\begin{figure*}[b]
  \centering
  \includegraphics[clip=true,width=\textwidth]{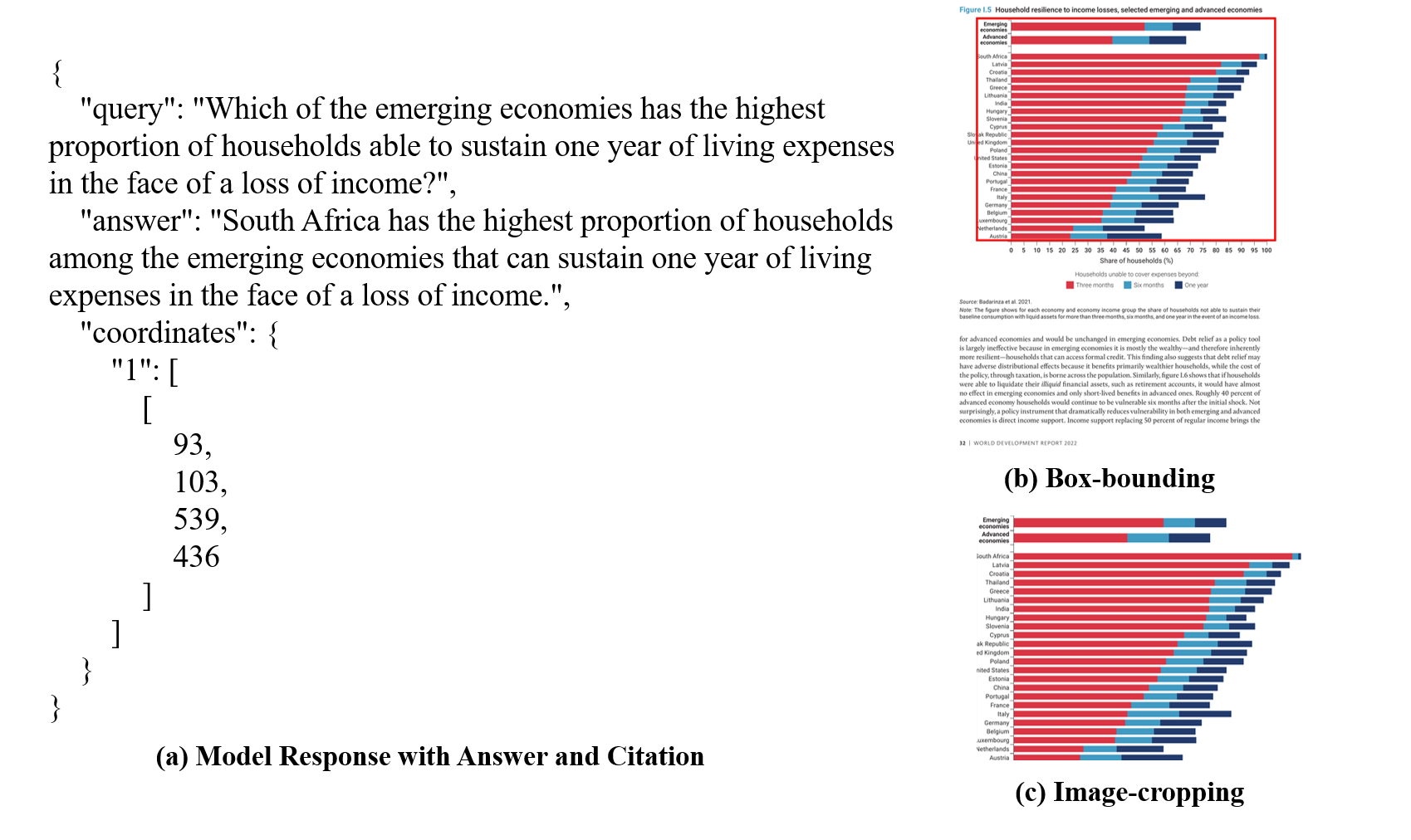}
  \caption{An example of MLLM's output of answer with visual citations, and the two evaluation methods: box-bounding and image-cropping. }
  \label{fig:box_crop}
\end{figure*}

Figure \ref{fig:box_crop} gives an example of the MLLM's output with both answer and citations, and demonstrates two citation evaluation methods: box-bounding and image-cropping.

\section{Case Study}\label{section:case study}
In this section, we provide several error cases based on both the different stages in the RGenCite baseline and the typical task types in finance.
\subsection{Error Case Study Based on Different Stages in RGenCite}
\begin{figure*}[h]
  \centering
  \includegraphics[clip=true,width=\textwidth]{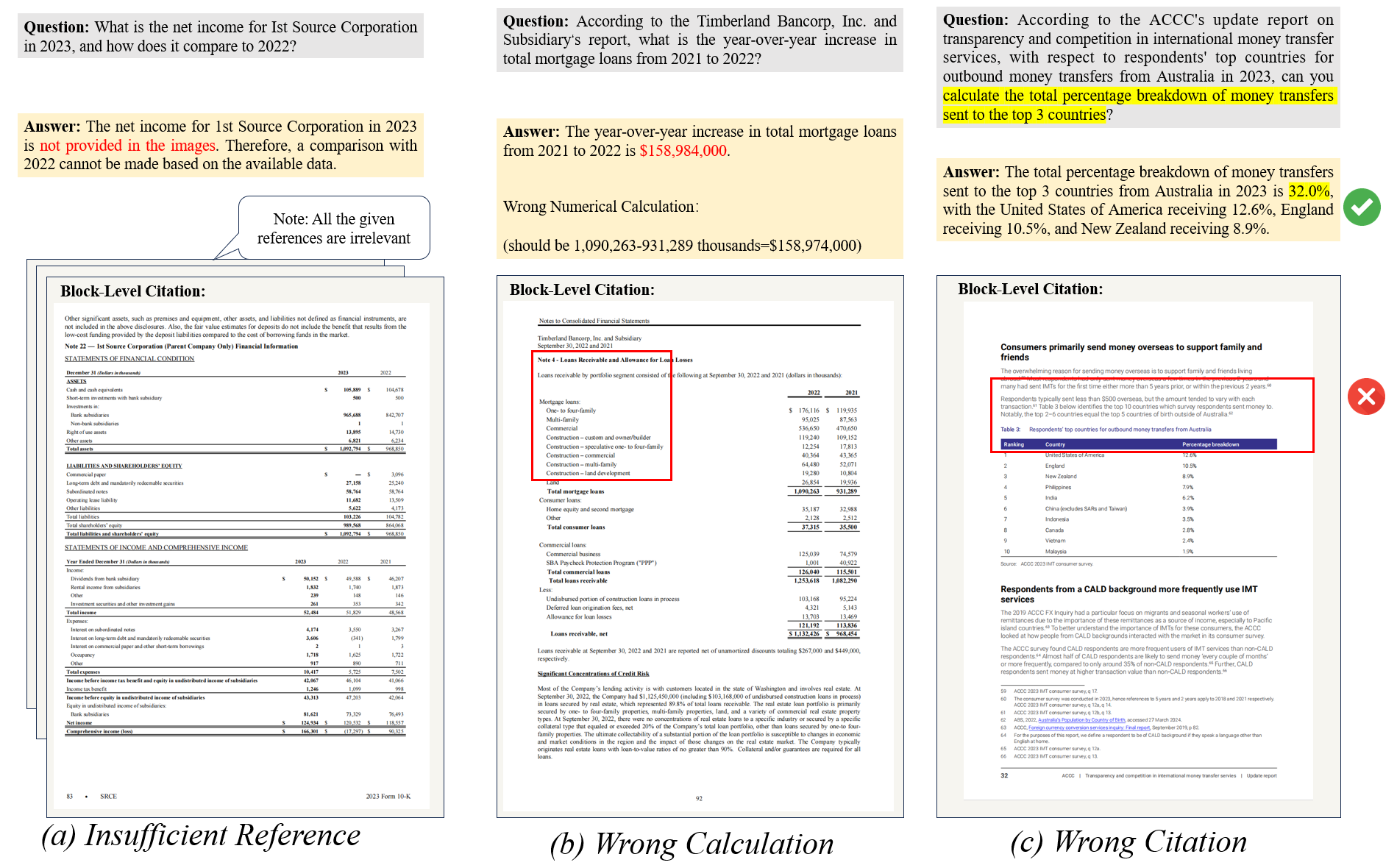}
  \caption{Three case study examples to illustrate the potential errors that can occur in RGenCite during generation and citation.}
  \label{fig:case_study}
\end{figure*}

To illustrate the potential errors that can occur in RGenCite during generation and citation, we conduct a case study identifying three main types of errors. The first type occurs when the retrieved reference image provided to the model lacks relevant information, resulting in insufficient data for the model to answer the question, as shown in Figure \ref{fig:case_study} (a). The second type involves providing the correct image, but the model makes an error in graphical reasoning, often leading to incorrect numerical calculations, as shown in Figure \ref{fig:case_study} (b). The third type occurs when the model answers the question correctly but introduces bias or inaccuracies in the citation, leading to incorrect referencing, as shown in Figure \ref{fig:case_study} (c).

\subsection{Error Case Study Based on Typical Task Types in the Financial Domain}
\begin{figure*}[h]
  \centering
  \includegraphics[clip=true,width=\textwidth]{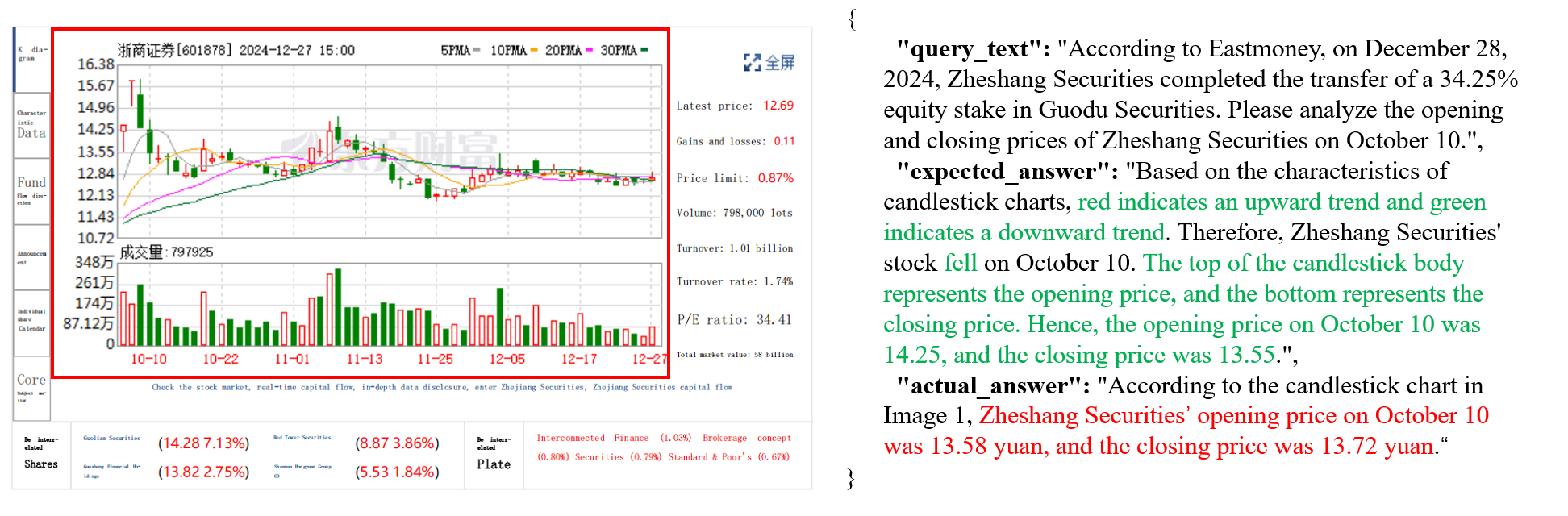}
  \caption{An Error Case of Information Extraction from Candlestick Chart}
  \label{fig:candlestick}
\end{figure*}

\paragraph{Recognizing Candlestick Charts.} As shown in Figure \ref{fig:candlestick}, for the query ``Based on the report from EastMoney, what are the opening and closing prices of Zheshang Securities on October 10, 2024?'' the correct analysis should recognize that red indicates an increase and green indicates a decrease in stock prices. The top of the candlestick body represents the opening price, while the bottom represents the closing price. In this case, the opening price was 14.25, and the closing price was 13.55. However, due to the lack of relevant knowledge, the models either produce incorrect results or generate responses like ``The image contains news reports about Zheshang Securities' acquisition of Guodu Securities shares and some securities market data, but it does not provide the specific opening and closing prices for Zheshang Securities on October 10, 2024''.

\begin{figure*}[h]
  \centering
  \includegraphics[clip=true,width=\textwidth]{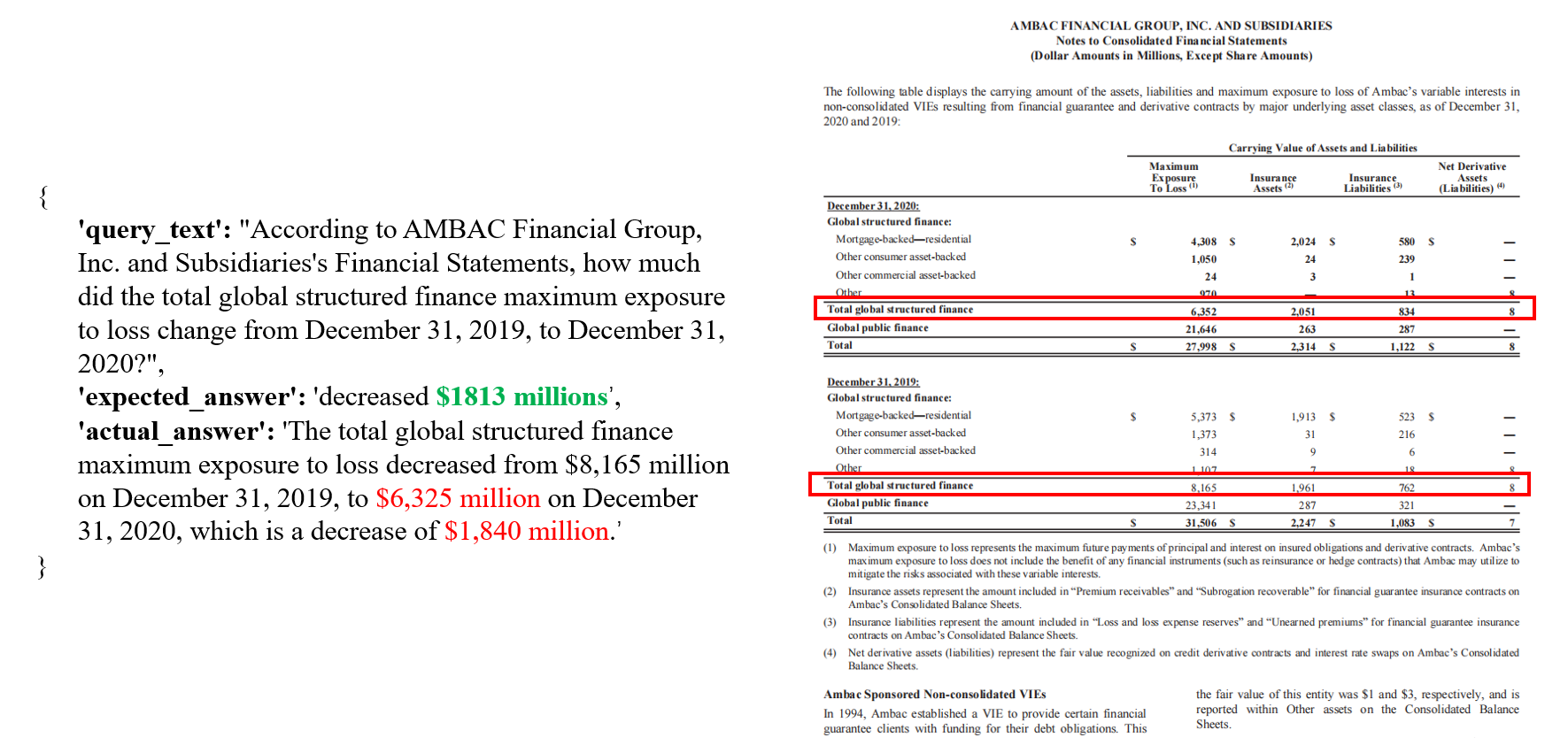}
  \caption{An Error Case of Numerical Calculation on Financial Table}
  \label{fig:calculate_case}
\end{figure*}

\paragraph{Dealing with Complex Financial Table.} Figure \ref{fig:calculate_case} is an error case that MLLMs fail in handling complex financial tables. In this case, the model was asked to calculate the change in total global structured finance maximum exposure to loss for AMBAC Financial Group, Inc. between December 31, 2019, and December 31, 2020. Although it correctly extracted the initial value of \$8,165 million, it mistakenly identified the ending value as \$6,325 million instead of the correct \$6,352 million. This minor misreading led to an incorrect computed decrease of \$1,840 million instead of the correct \$1,813 million. Such errors reveal the challenges MLLMs face in accurately interpreting numeric details from financial tables, where even small misreads can lead to significant factual inaccuracies.

\begin{figure*}[h]
  \centering \includegraphics[clip=true,width=\textwidth]{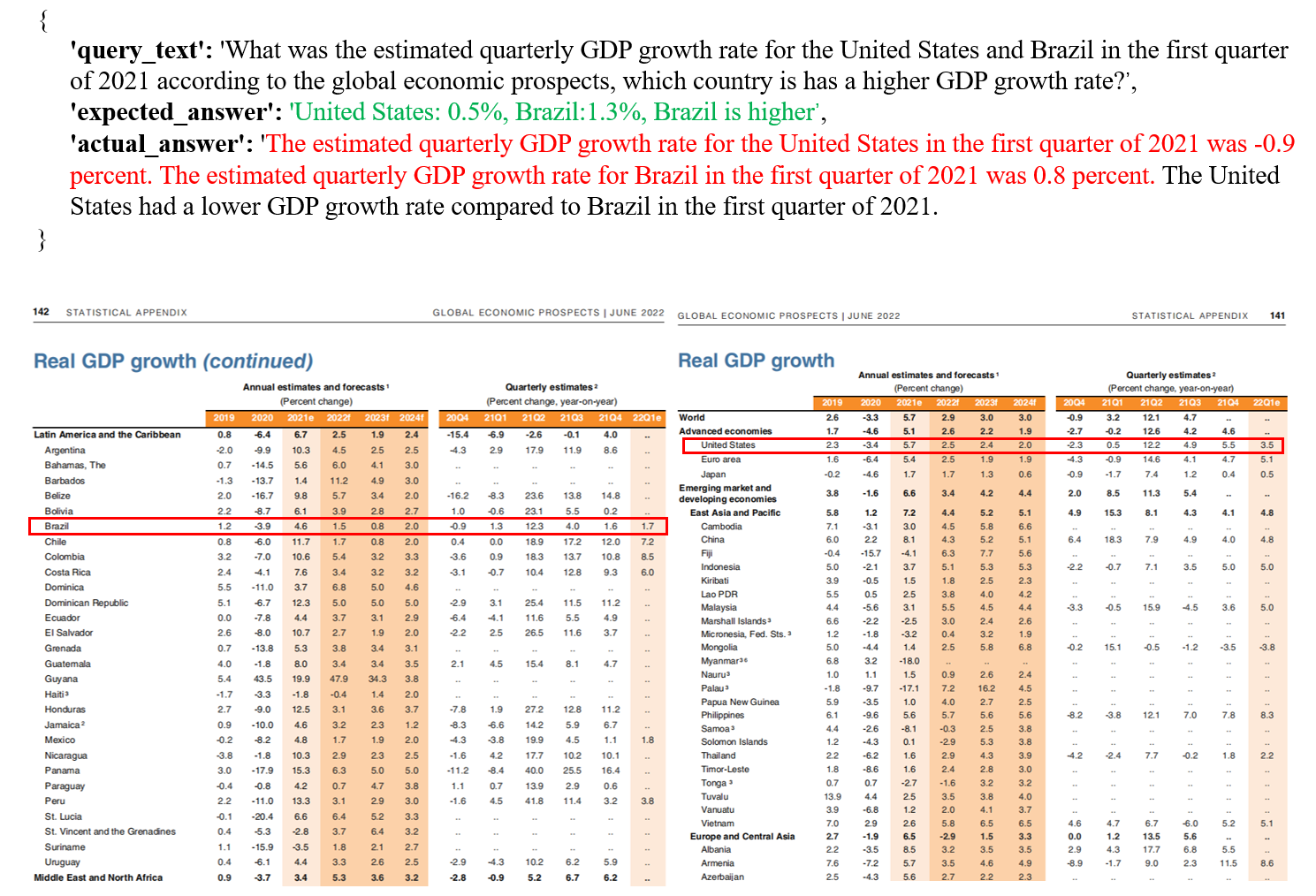}
  \caption{An Error Case of Multi-page Question}
  \label{fig:case_multipage}
\end{figure*}

\paragraph{Dealing with Multi-page Questions.} The example in Figure \ref{fig:case_multipage} illustrates a typical limitation of MLLMs when dealing with lengthy financial tables that span multiple pages. The model was asked to extract and compare the quarterly GDP growth rates for the United States and Brazil in Q1 2021 from the Global Economic Prospects report. However, the relevant data was distributed across two separate pages, and the model failed to aggregate the information correctly. As a result, it misreporting the growth rate of Brazil and the U.S., leading to an inaccurate comparison. This case highlights the difficulty MLLMs face in maintaining contextual continuity across paginated tables, a common format in financial documents.

\section{Annotation guidelines.}\label{sec:guideline}
This section demonstrates the annotation guidelines. The annotation guidelines for constructing the QA dataset in shown Table \ref{tab:qa_annotation_guideline} and the guideline for rating-based human evaluation for visual citation is in Table \ref{tab:citation_annotation_guideline}.

\section{Resource Usage}\label{sec:resource_usage}
Throughout the processes of dataset construction, response generation, and evaluation, we employed multiple proprietary language model APIs, including GPT-4o and other commercial multimodal large language models (MLLMs). The total API usage cost amounted to \$3,021.47. All experiments with open-source models were conducted locally on 4×A100 80GB GPUs. The dataset was manually annotated by three experienced annotators to ensure quality and consistency.

We relied on several mainstream libraries and toolkits across retrieval, generation, and evaluation tasks, including PyTorch, Transformers, pytrec\_eval, pylate.

We carefully considered the licenses and intended use cases of all third-party artifacts utilized in our study. All datasets and tools used from external sources were employed strictly within the bounds of their respective licenses and intended purposes, primarily for academic research. 

\section{Potential Risks}
Despite careful design and construction, our retrieval corpus and QA dataset may still contain potential risks. During the data collection process, some noisy, outdated, or irrelevant financial documents might not have been fully filtered. Similarly, in the QA dataset, there may be annotation errors, ambiguities, or biases due to imperfect filtering and manual oversight. These issues could affect the accuracy of model evaluation and the generalizability of experimental results. We encourage users of FinRAGBench-V to be aware of these limitations and apply additional validation where necessary.

\begin{table*}[htbp]
    \footnotesize
    \centering
    \begin{tabular}{p{14cm}}
        \toprule
        \textbf{Instruction:} Here is an image of a document. Your task is to generate queries about this document image from various perspectives, categorize the questions (category), provide answers to the questions (answer), and specify whether the answer is a long or short answer (answer\_type). \\
        \#\#\#I hope your questions are as detailed as possible. Begin by specific about which document you are referring to and describe the required text, table, or chart content without explicitly mentioning the figure or table number. \\
        \#\#\#Your questions can target the text, tables, charts, or any other elements in the image. \\
        \#\#\#Design three different queries for each document, ensuring that the question categories (category) are distinct from each other. \\
        \#\#\#The categories of questions you can include are:
        Text-based QA:\\
        1. Text-Text Inference: Extraction or reasoning based on textual information.\\\\
        
        Chart-based QA:\\
        1. Chart-Information Extraction: Extract key metrics or features from the chart.\\
        2. Chart-Numerical Calculation: Includes calculations such as growth rates, interest rates, total costs, etc.\\
        3. Chart-Time-Sensitive: Includes trend descriptions, causal relationships, event sequences, frequencies, durations, etc.\\\\
        
        Table-based QA:\\
        1. Table-Numerical Calculation: Perform calculations such as growth rates, interest rates, total costs, etc., using table data.\\
        2. Table-Comparison and Sorting: Compare or rank entities based on specific criteria (e.g., return rates, risks).\\\\
        
        Here is the format of your output: \\
        \begin{verbatim}
        {
            "result":[
                {
                    "query" : "",
                    "category":"",
                    "answer": "",
                    "answer_type":""
                },
                {
                    "answer": "",
                    "query" : "",
                    "category":"",
                    "answer_type":""
                },
                {
                    "answer": "",
                    "query" : "",
                    "category":"",
                    "answer_type":""
                }
            ]
        }
        \end{verbatim} \\
        Here are some examples:\\
         \{examples\}\\
        \bottomrule
    \end{tabular}
    \caption{Prompt for Constructing QA Dataset}
    \label{tab:Prompt for Constructing QA Dataset}
\end{table*}

\begin{table*}[htbp]
    \footnotesize
    \centering
    \begin{tabular}{p{14cm}}
        \toprule
        \textbf{Instruction:} Answer the following questions based on the given images, identify the images that support your answer, and further locate the source of your answer in the images by outputting coordinate pairs. \\
        \#\#\#If the answer uses more than one image, you must point out all the images used; If your answer uses information from more than one image, you must annotate all the used information. \\
        \#\#\#All your annotations must fully support your answer, and there must not be any unsupported information in your answer. \\
        \#\#\#When annotating an image, you need to annotate a full graph or text paragraph, not just a specific number. \\
        Your replies must strictly follow the following JSON format: \\
        \begin{verbatim}
        {
            "answer":"",
            "coordinates":{
            "1":[[x1, y1, x2, y2], [x1, y1, x2, y2]],
            "2":[[x1, y1, x2, y2], [x1, y1, x2, y2]],
                ... # These are the supportive images and the coordinate pairs in them
            }
        }
        \end{verbatim} \\
        \textbf{Here is the question:} \{query\}\\
        \textbf{Here are the images:} \\
        Image 1: Width: {width1}, Height: {height1} \\
        (Image 1 in Base64) \\
        Image 2: Width: {width2}, Height: {height2} \\
        (Image 2 in Base64) \\
        . \\
        . \\
        . \\
        \bottomrule
    \end{tabular}
    \caption{Prompt for Generation and Citation}
    \label{tab:Prompt for Generation and Citation}
\end{table*}

\begin{table*}[htbp]
    \footnotesize
    \centering
    \begin{tabular}{p{14cm}}
        \toprule
        \textbf{Question:} \{query\_text\} \\
        \textbf{Ground\_truth:} \{expected\_answer\} \\
        \textbf{Model\_answer:} \{actual\_answer\} \\
        Is the model answer correct? You only need to output ‘true’ for correct or ‘false’ for incorrect. If the model answer does not contain any information, it should be judged as ‘false’. \\
        \bottomrule
    \end{tabular}
    \caption{Prompt for Response Accuracy Evaluation}
    \label{tab:acc_prompt}
\end{table*}

\begin{table*}[htbp]
    \footnotesize
    \centering
    \begin{tabular}{p{14cm}}
        \toprule
        \textbf{Answer:} \{answer\}
        Please judge whether these pages cover the answer, your answer can only be 'yes' or 'no'. \\
        \textbf{Here are my images:}\\
        (Image 1 in Base64)\\
        (Image 2 in Base64)\\
        . \\
        . \\
        . \\
        \bottomrule
    \end{tabular}
    \caption{Prompt for Page-Level Citation Evaluation}
    \label{tab:citation_prompt_page}
\end{table*}

\begin{table*}[htbp]
    \footnotesize
    \centering
    \begin{tabular}{p{14cm}}
        \toprule
        \textbf{Answer:} \{answer\}
        The following images will contain marked areas (red boxes), please judge whether these marked areas (red boxes) cover the content of the answer, your answer can only be 'yes' if it covers or 'no' if it doesn't cover. \\
        \textbf{Here are my images:}\\
        (Image 1 in Base64)\\
        (Image 2 in Base64)\\
        . \\
        . \\
        . \\
        \bottomrule
    \end{tabular}
    \caption{Prompt for Block-Level Citation Evaluation using Box-Bounding}
    \label{tab:citation_prompt_bounding_boxes}
\end{table*}

\begin{table*}[htbp]
    \footnotesize
    \centering
    \begin{tabular}{p{14cm}}
    \toprule
        \textbf{Answer:} \{answer\}
        Below are some extracts from the images, please decide if they cover the answers given, your answer can only be 'yes' if it covers or 'no' if it doesn't cover. \\
        \textbf{Here are my images:}\\
        (Image 1 in Base64)\\
        (Image 2 in Base64)\\
        . \\
        . \\
        . \\
    \bottomrule
    \end{tabular}
    \caption{Prompt for Block-Level Citation Evaluation using Image-Cropping}
    \label{tab:citation_prompt_image_cropping}
\end{table*}

\begin{table*}[h!]
\centering
\begin{tabular}{lp{10cm}} 
    \toprule
    \textbf{Query:} &In Howden Joinery Group Plc's Annual Report \& Accounts 2022, with respect to the Nominations Committee report for 2022, who is mentioned as the individual appointed to lead the Committee and who retired?\\
    \textbf{Category:} & Text Inference \\
    \textbf{Answer:} &  Peter Ventress was appointed as the Committee Chairman, and Richard Pennycook retired. \\
    \textbf{Reference Image:} & \\
    &\includegraphics[width=0.6\textwidth]{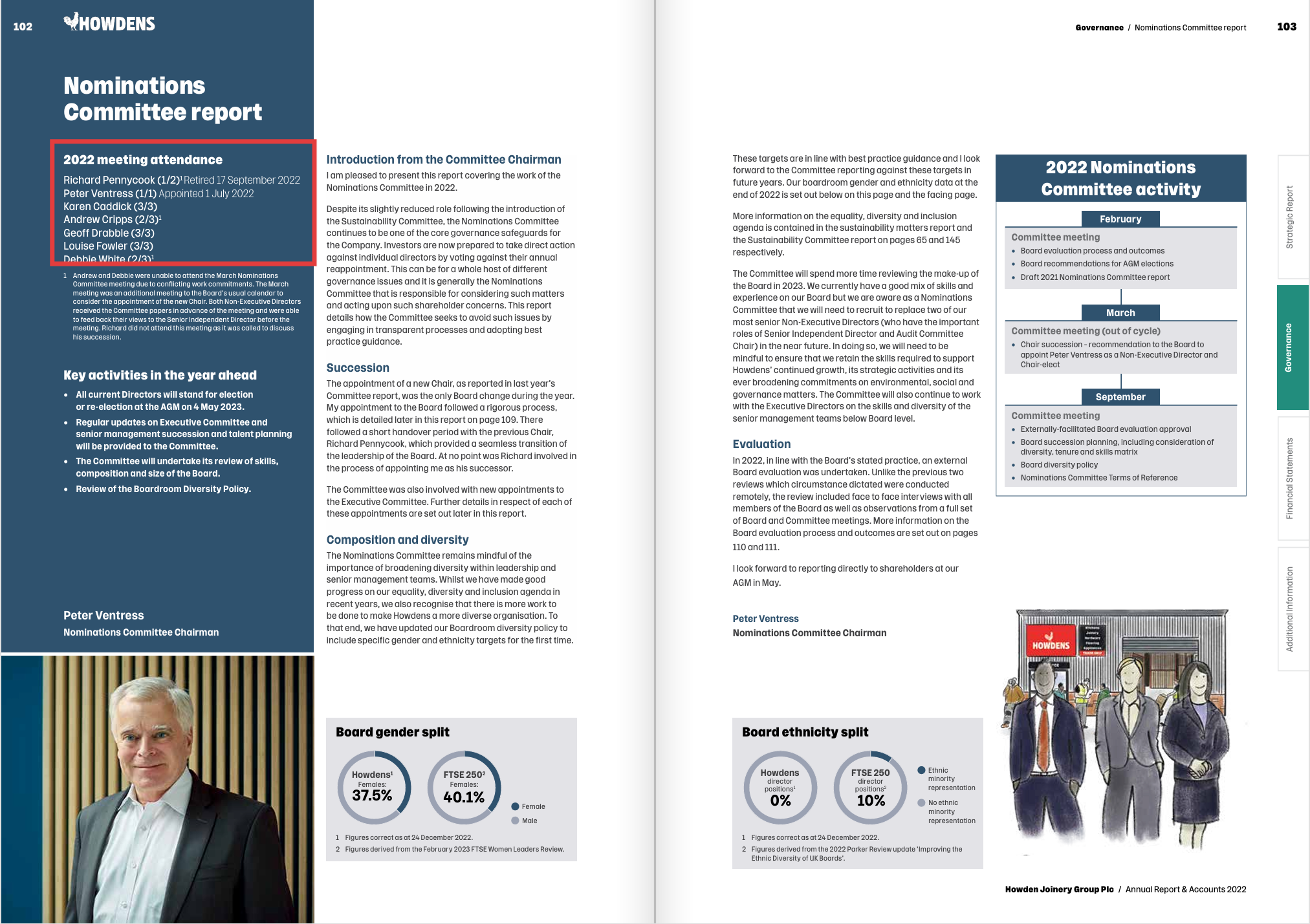} \\
    \bottomrule
\end{tabular}
\caption{QA Dataset Example 1: An Example of Text Inference Question}
\label{tab:text inference}
\end{table*}

\begin{table*}[h!]
\centering
\begin{tabular}{lp{10cm}} 
    \toprule
    \textbf{Query:} &From the document 'Independent auditors' report to the members of Craneware plc', what is the significance of revenue recognition as a key audit matter in the context of the Group's financial statement?\\
    \textbf{Category:} & Text Inference \\
    \textbf{Answer:} & Revenue recognition is significant because it involves determining the amount of revenue to be recognized based on contract details and conditions in contracts with customers. The risk is identified at the journal level related to the existence and occurrence of all revenue streams. \\
    \textbf{Reference Image:} & \\
    &\includegraphics[width=0.6\textwidth]{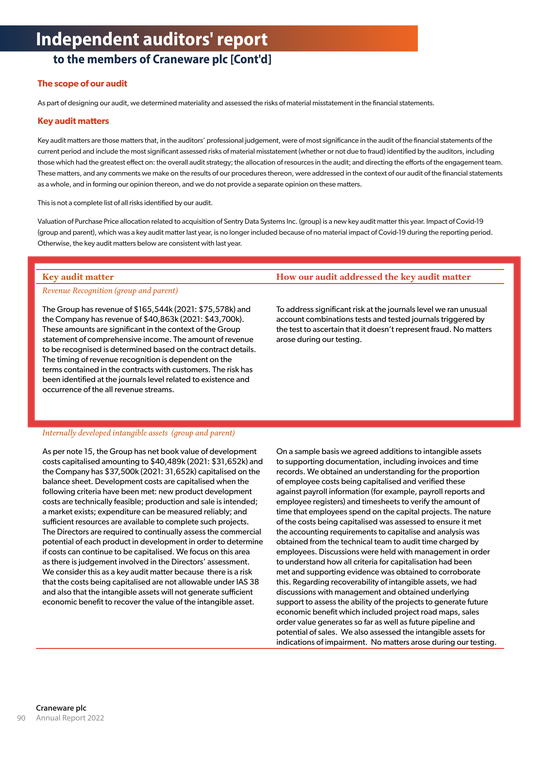} \\
    \bottomrule
\end{tabular}
\caption{QA Dataset Example 2: An Example of Text Inference Question}
\label{tab:text inference2}
\end{table*}

\begin{table*}[h!]
\centering
\begin{tabular}{lp{10cm}} 
    \toprule
    \textbf{Query:} & According to the Annual Report and Account for Howden Joinery Group Plc in 2023, what is the total baseline emissions estimation for 2021? How many percentage does the purchased goods and services take among them? \\
    \textbf{Category:} & Chart-Information Extraction \\
    \textbf{Answer:} & The total 2021 baseline emissions are estimated at 1.2m \ch{\{TCO_2e\}}. Among them, purchased goods and services takes 40\%. \\
    \textbf{Reference Image:} & \\
    &\includegraphics[width=0.6\textwidth]{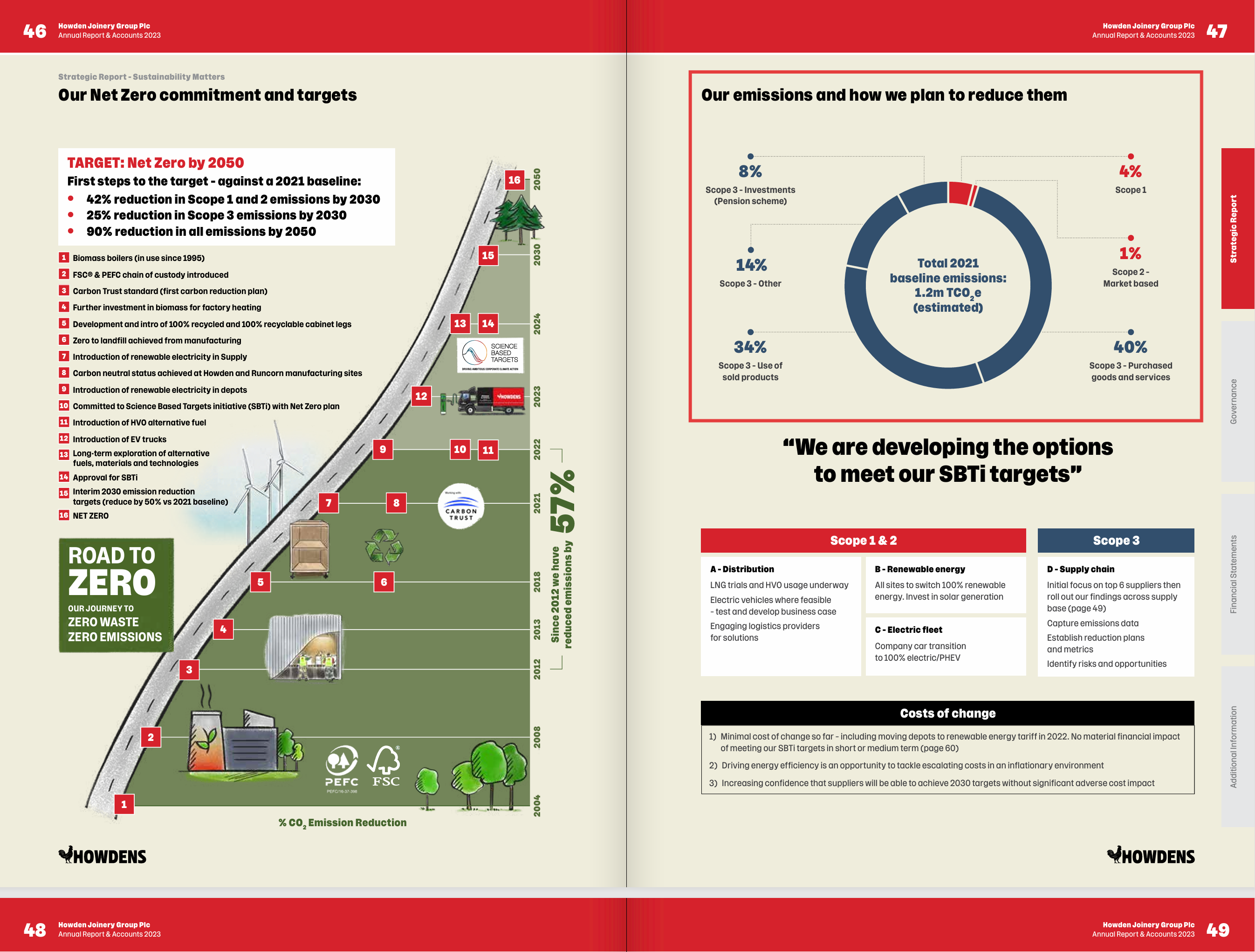} \\
    \bottomrule
\end{tabular}
\caption{QA Dataset Example 3: An Example of Chart-Information Exraction Question}
\label{tab:chart info extract}
\end{table*}

\begin{table*}[h!]
\centering
\begin{tabular}{lp{10cm}} 
    \toprule
    \textbf{Query:} & According to IFC's 2024 annual report, among all the IFC's funding resources, which one is the highest?\\
    \textbf{Category:} & Chart-Information Extraction \\
    \textbf{Answer:} & Borrowings from market resources. \\
    \textbf{Reference Image:} & \\
    &\includegraphics[width=0.6\textwidth]{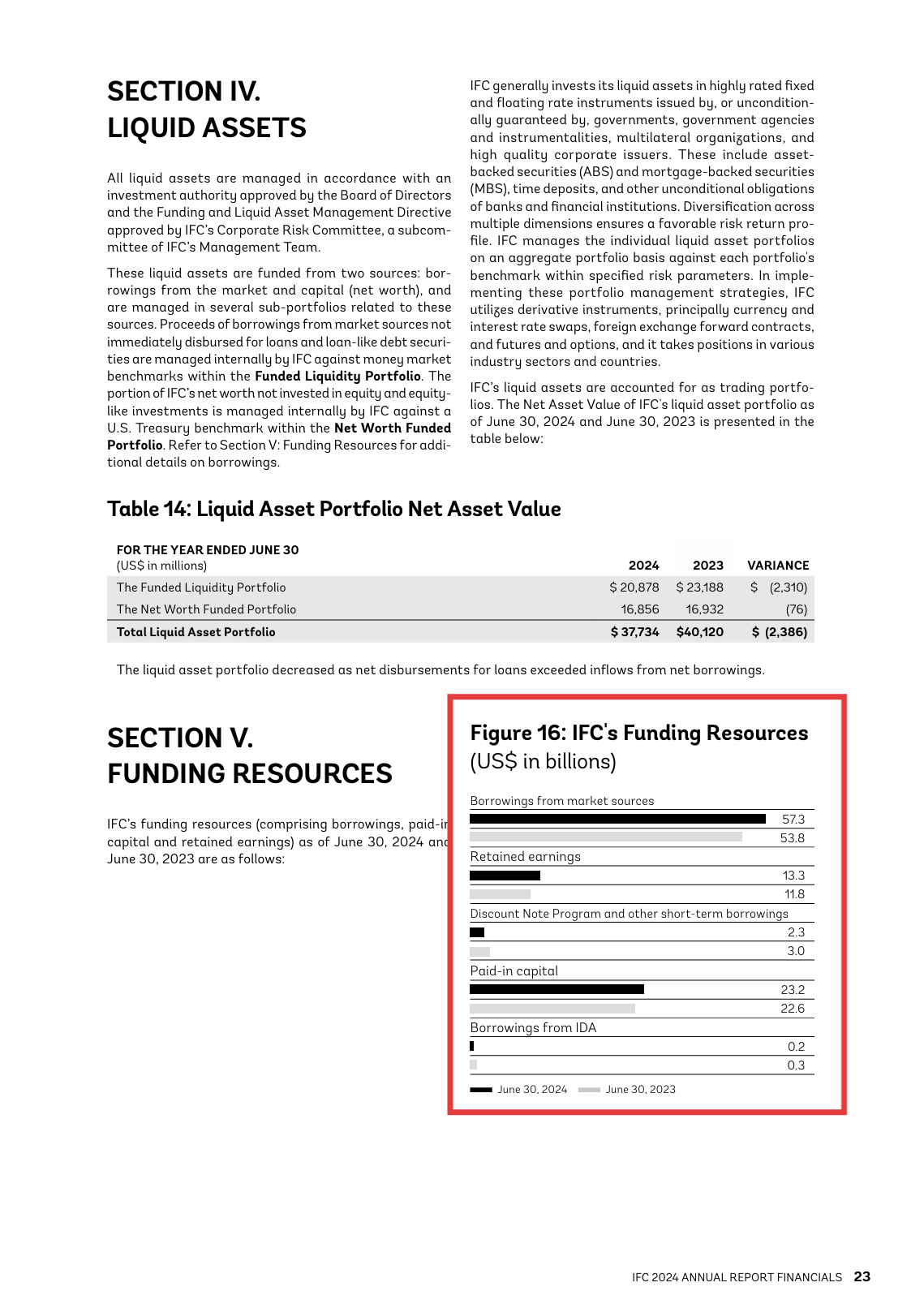} \\
    \bottomrule
\end{tabular}
\caption{QA Dataset Example 4: An Example of Chart-Information Exraction Question}
\label{tab:chart info extract2}
\end{table*}

\begin{table*}[h!]
\centering
\begin{tabular}{lp{10cm}} 
    \toprule
    \textbf{Query:} &Analyzing the Private Financing Deal Count reported by FinTech Insights in Q3 2024, how many financing deals did it increased from Q1 2021 to Q2 2021? \\
    \textbf{Category:} & Chart-Numerical Calculations \\
    \textbf{Answer:} & 18 \\
    \textbf{Reference Image:} & \\
    &\includegraphics[width=0.6\textwidth]{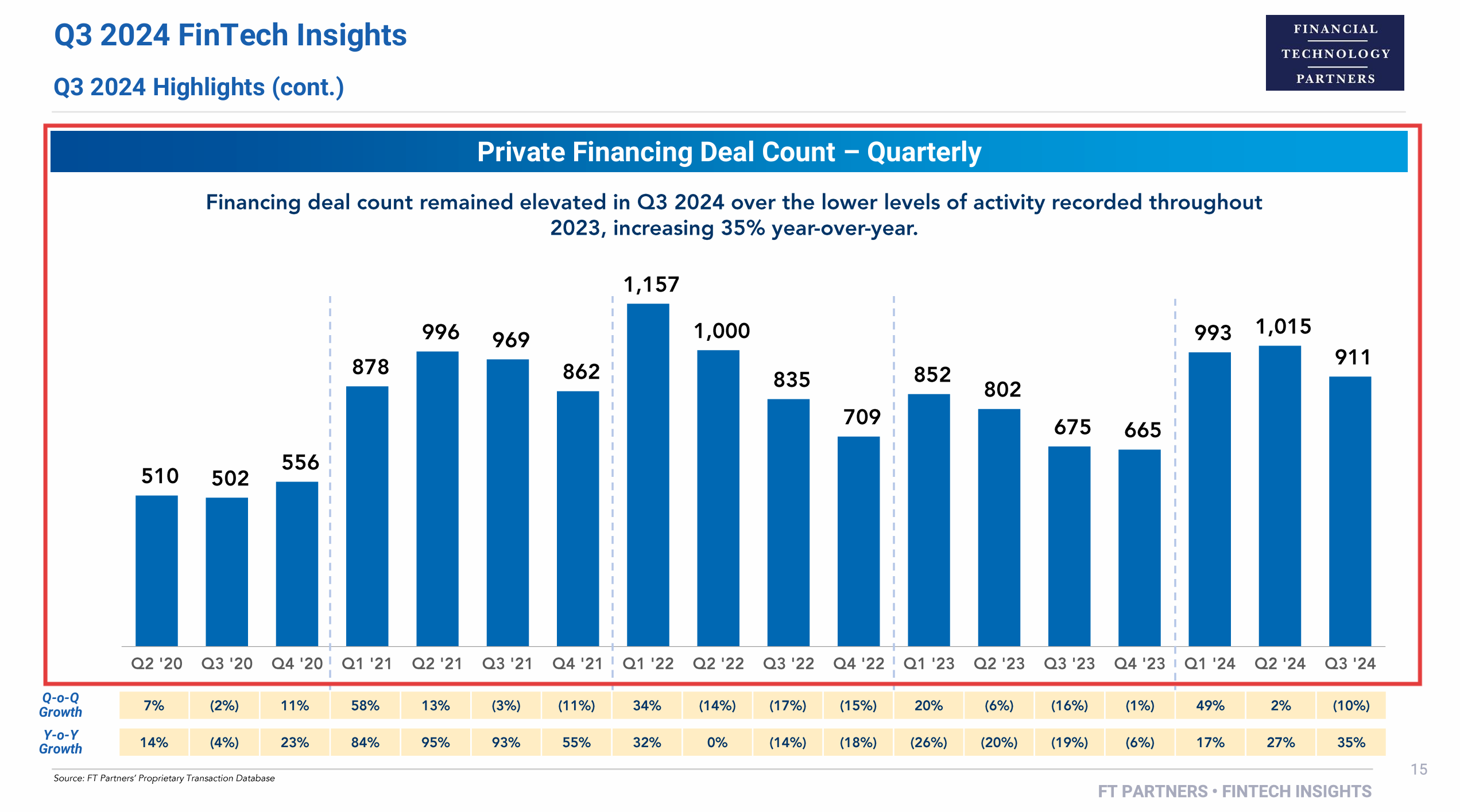} \\
    \bottomrule
\end{tabular}
\caption{QA Dataset Example 5: An Example of Chart-Numerical Calculations Question}
\label{tab:chart numerical}
\end{table*}

\begin{table*}[h!]
\centering
\begin{tabular}{lp{10cm}} 
    \toprule
    \textbf{Query:} &Based on the statistics of climate finance flows by international and domestic, what is the growth rate of domestic public funding from 2019/20 to 2021/22?\\
    \textbf{Category:} & Chart-Numerical Calculations \\
    \textbf{Answer:} & -37.5\% \\
    \textbf{Reference Image:} & \\
    &\includegraphics[width=0.6\textwidth]{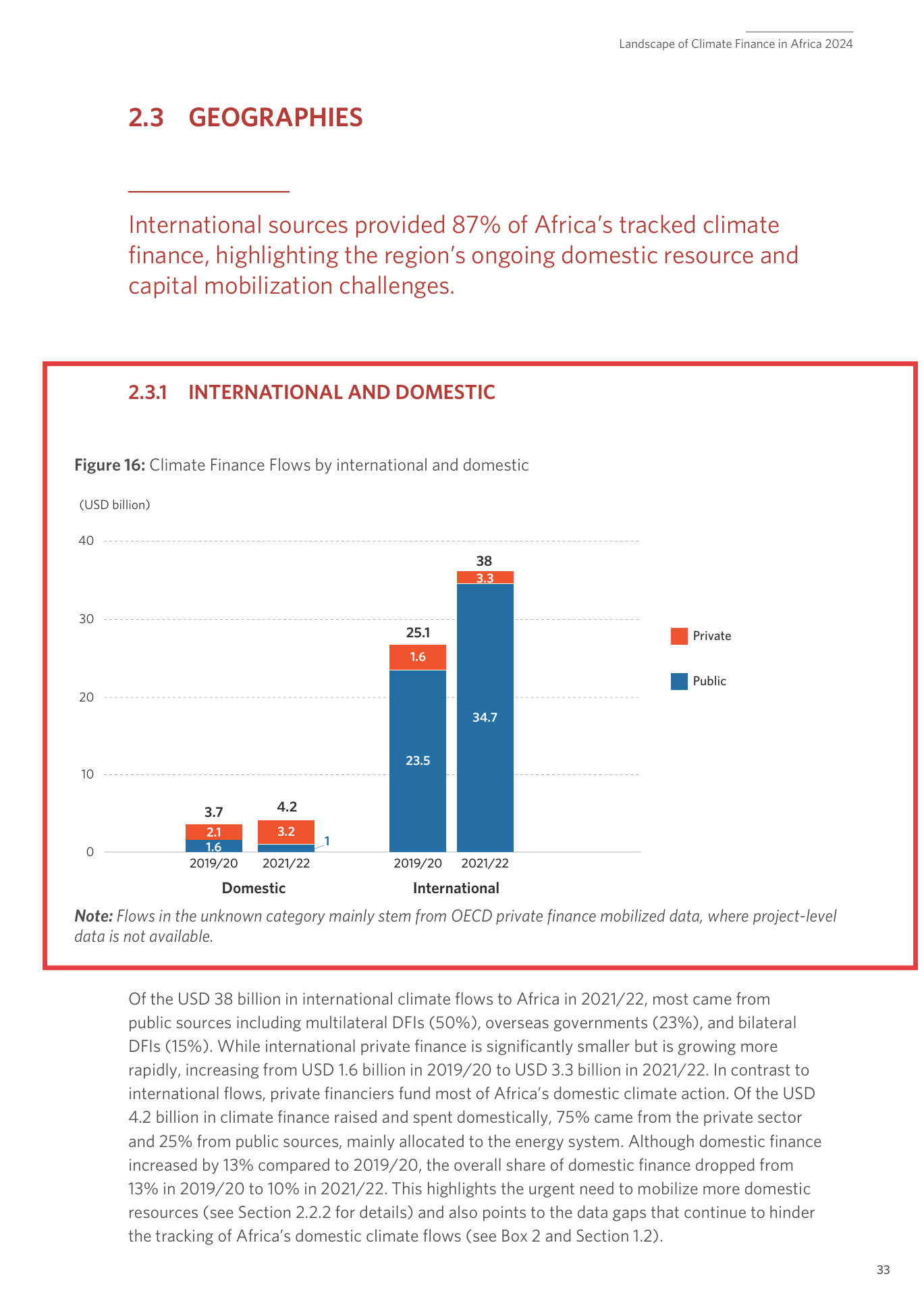} \\
    \bottomrule
\end{tabular}
\caption{QA Dataset Example 6: An Example of Chart-Numerical Calculations Question}
\label{tab:chart numerical2}
\end{table*}

\begin{table*}[h!]
\centering
\begin{tabular}{lp{10cm}} 
    \toprule
    \textbf{Query:} &According to Howden Joinery Group Plc Annual Report \& Accounts 2021, what is the trend of depot openings in the UK and France from 2017 to 2021? \\
    \textbf{Category:} & Chart-Time Sensitive \\
    \textbf{Answer:} & There's a consistent increase in depot openings from 2017 to 2021, with a particularly significant increase in 2021. \\
    \textbf{Reference Image:} & \\
    &\includegraphics[width=0.6\textwidth]{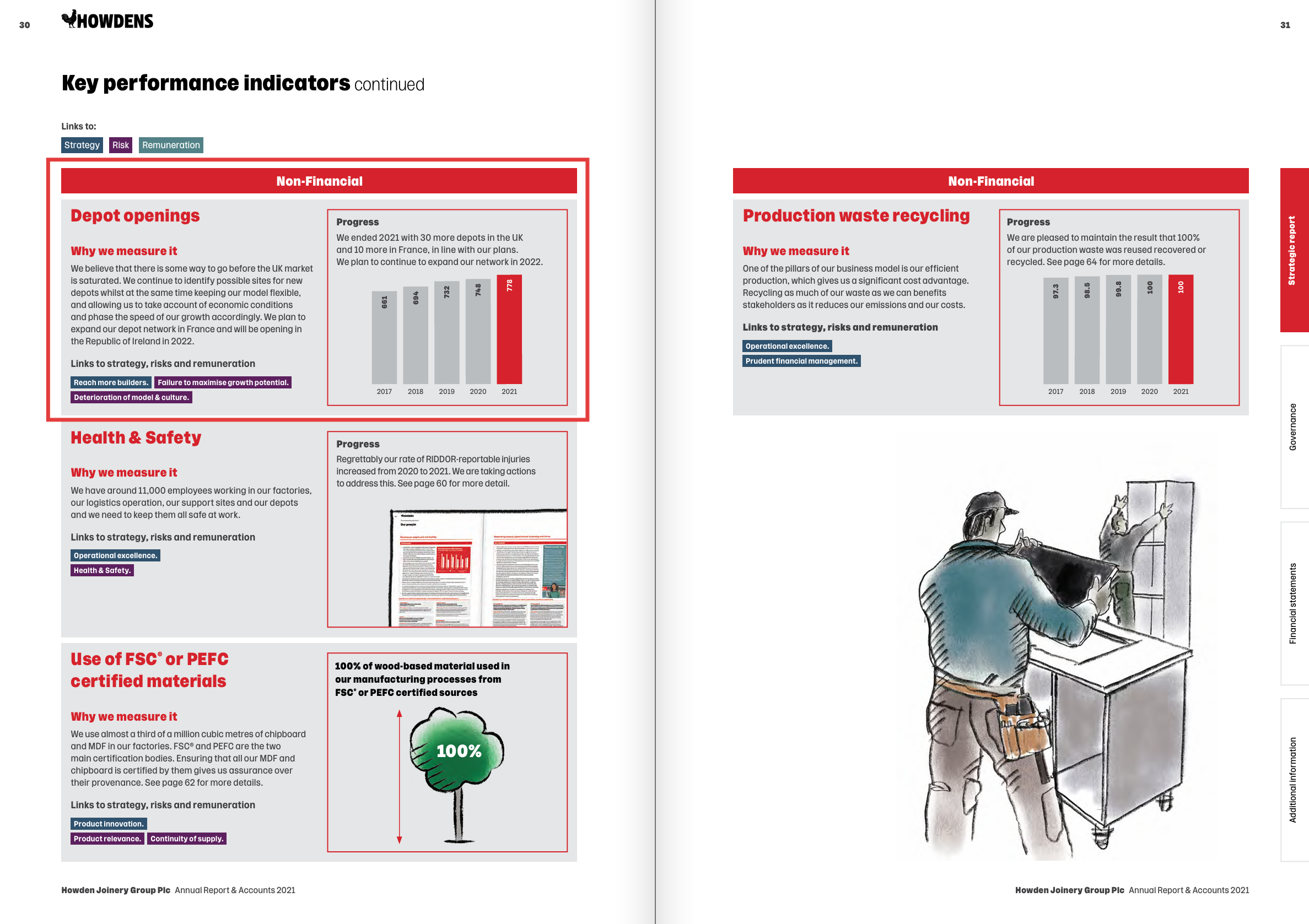} \\
    \bottomrule
\end{tabular}
\caption{QA Dataset Example 7: An Example of Chart-Time Sensitive Question}
\label{tab:chart time}
\end{table*}

\begin{table*}[h!]
\centering
\begin{tabular}{lp{10cm}} 
    \toprule
    \textbf{Query:} &According to the Wall Street stocks data from July 31,2024 to Aug 13,2024, explain the trends of S\&P 500 and Nasdaq Composite indices during that time period.\\
    \textbf{Category:} & Chart-Time Sensitive \\
    \textbf{Answer:} & There's a steep decline followed by a bounce back for both the S\&P 500 and Nasdaq Composite indices. After an initial drop where both indices reached close to their lowest points, they recovered steadily with the Nasdaq Composite seeing a slightly stronger recovery than the S\&P 500. This indicates a volatile period followed by a short-term rebound. \\
    \textbf{Reference Image:} & \\
    &\includegraphics[width=0.6\textwidth]{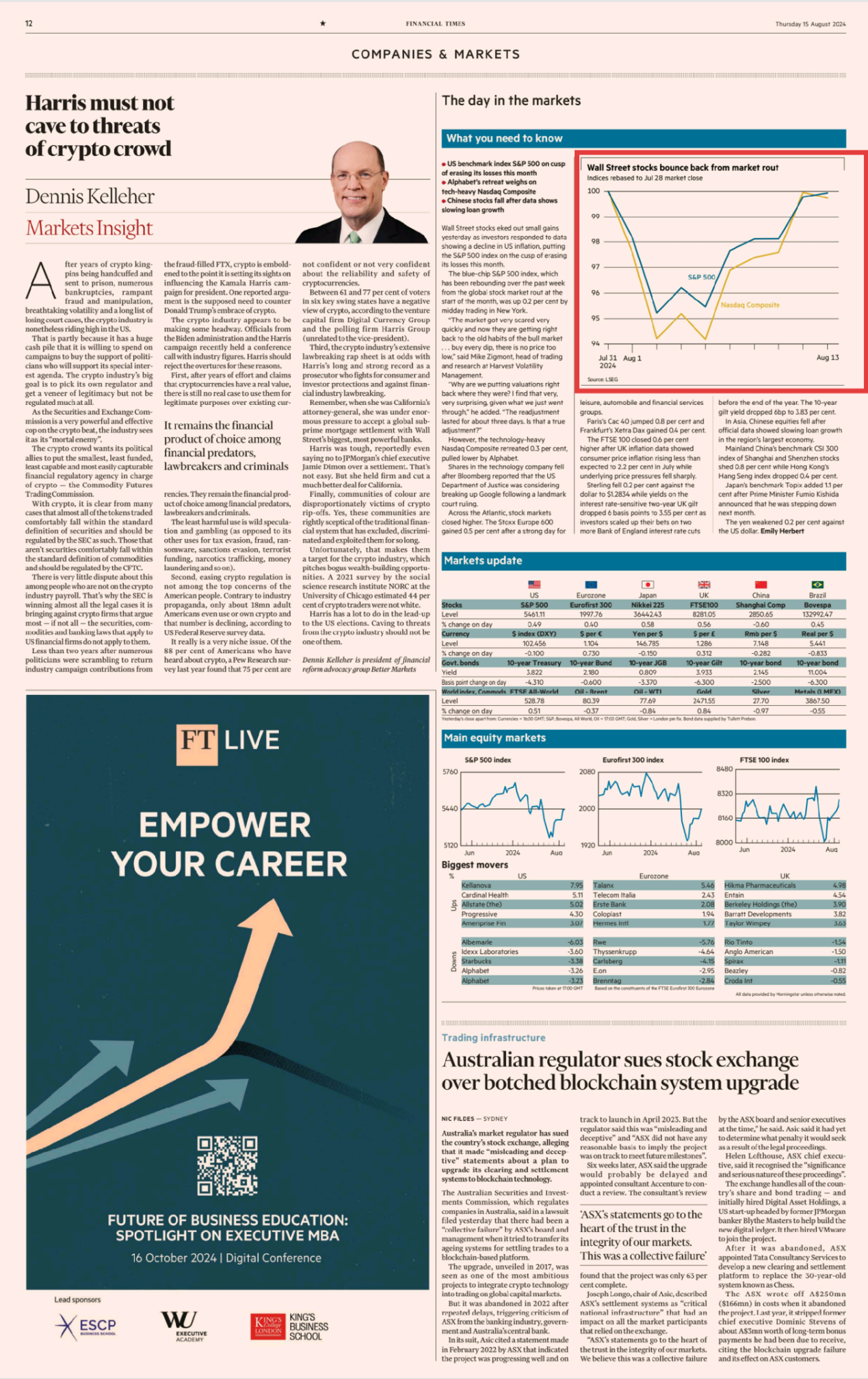} \\
    \bottomrule
\end{tabular}
\caption{QA Dataset Example 8: An Example of Chart-Time Sensitive Question}
\label{tab:chart time2}
\end{table*}

\begin{table*}[h!]
\centering
\begin{tabular}{lp{10cm}} 
    \toprule
    \textbf{Query:} &Based on the data under the 'Related party transactions' in the Craneware plc Annual Report and Financial Statements 2023, what is the percent increase in Salaries and short-term employee benefits for Executive Directors from 2022 to 2023? \\
    \textbf{Category:} & Table-Numerical Calculations \\
    \textbf{Answer:} & An increase of approximately 84.94\%. \\
    \textbf{Reference Image:} & \\
    &\includegraphics[width=0.6\textwidth]{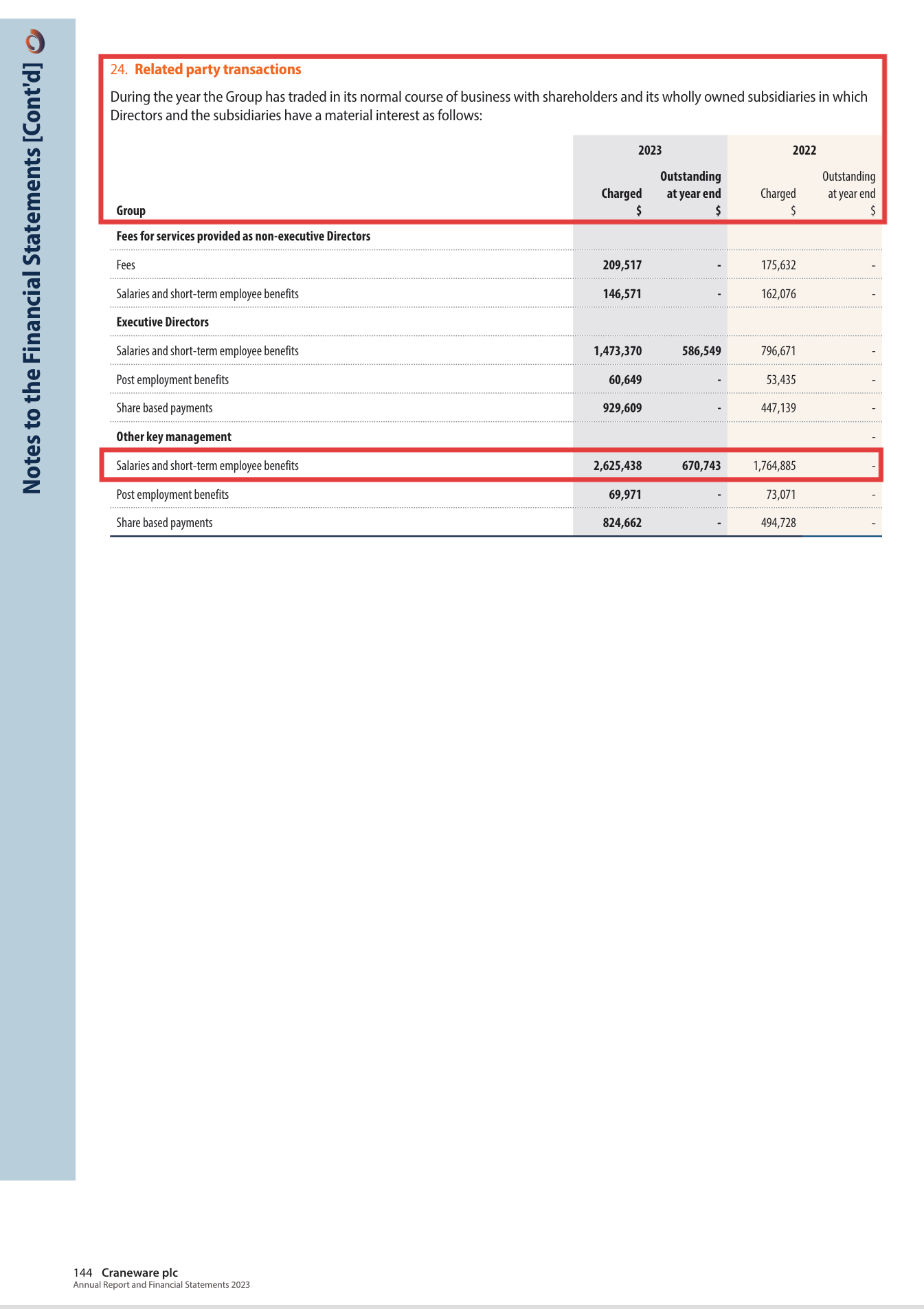} \\
    \bottomrule
\end{tabular}
\caption{QA Dataset Example 9: An Example of Table-Numerical Calculations Question}
\label{tab:table-numerical}
\end{table*}

\begin{table*}[h!]
\centering
\begin{tabular}{lp{10cm}} 
    \toprule
    \textbf{Query:} &According to the Q3 2024 FinTech Insights document, with respect to Publicly Traded FinTech Companies – Selected Top Performers in 2024 YTD, what is the combined H1 2024 Return for all companies categorized under 'InsurTech'?\\
    \textbf{Category:} & Table-Numerical Calculations \\
    \textbf{Answer:} &The combined H1 2024 Return for companies under 'InsurTech' is 449\%. This is calculated by adding the returns of Root Insurance (260\%), Hippo (85\%), and Policybazaar.com (104\%). \\
    \textbf{Reference Image:} & \\
    &\includegraphics[width=0.6\textwidth]{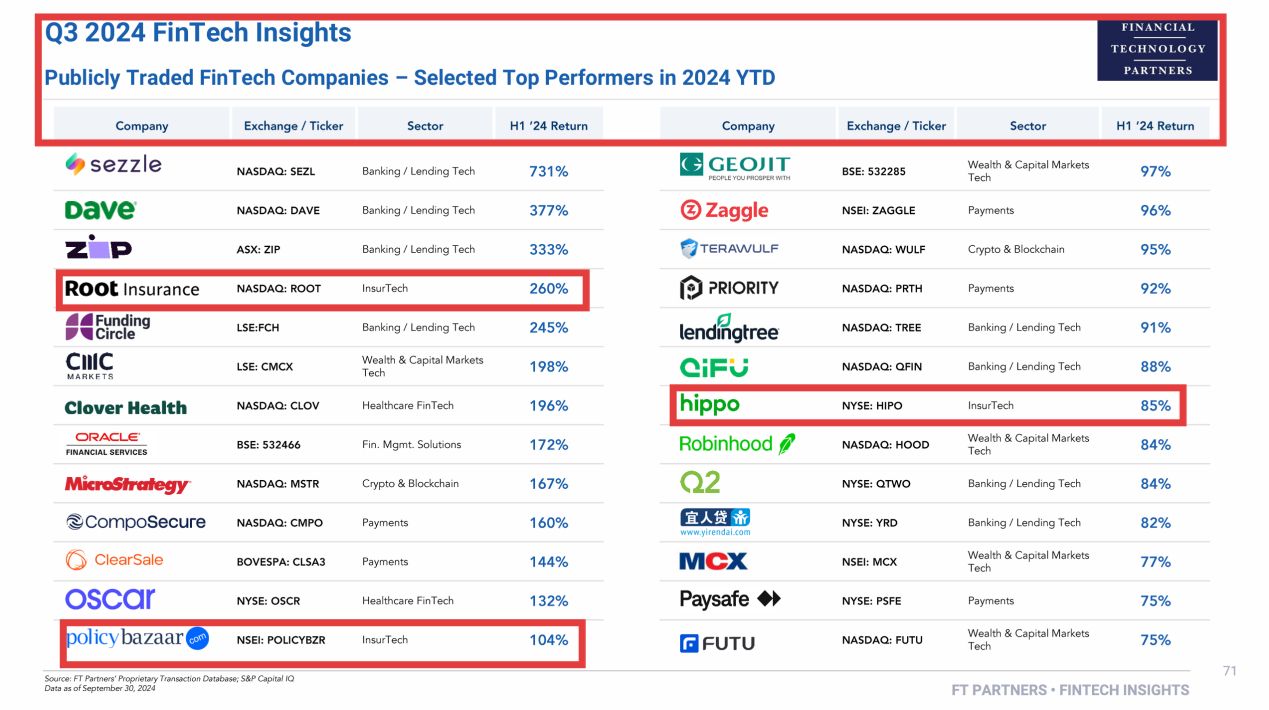} \\
    \bottomrule
\end{tabular}
\caption{QA Dataset Example 10: An Example of Table-Numerical Calculations Question}
\label{tab:table-numerical2}
\end{table*}

\begin{table*}[h!]
\centering
\begin{tabular}{lp{10cm}} 
    \toprule
    \textbf{Query:} &According to the 2022 annual report of Craneware plc, which plan had the larger exercise price range: the 2016 Schedule 4 Option Plan or the 2018 SAYE Option Plan? \\
    \textbf{Category:} &Table-Comparison and Sorting \\
    \textbf{Answer:} &2016 Schedule 4 Option Plan.\\
    \textbf{Reference Image:} & \\
    &\includegraphics[width=0.6\textwidth]{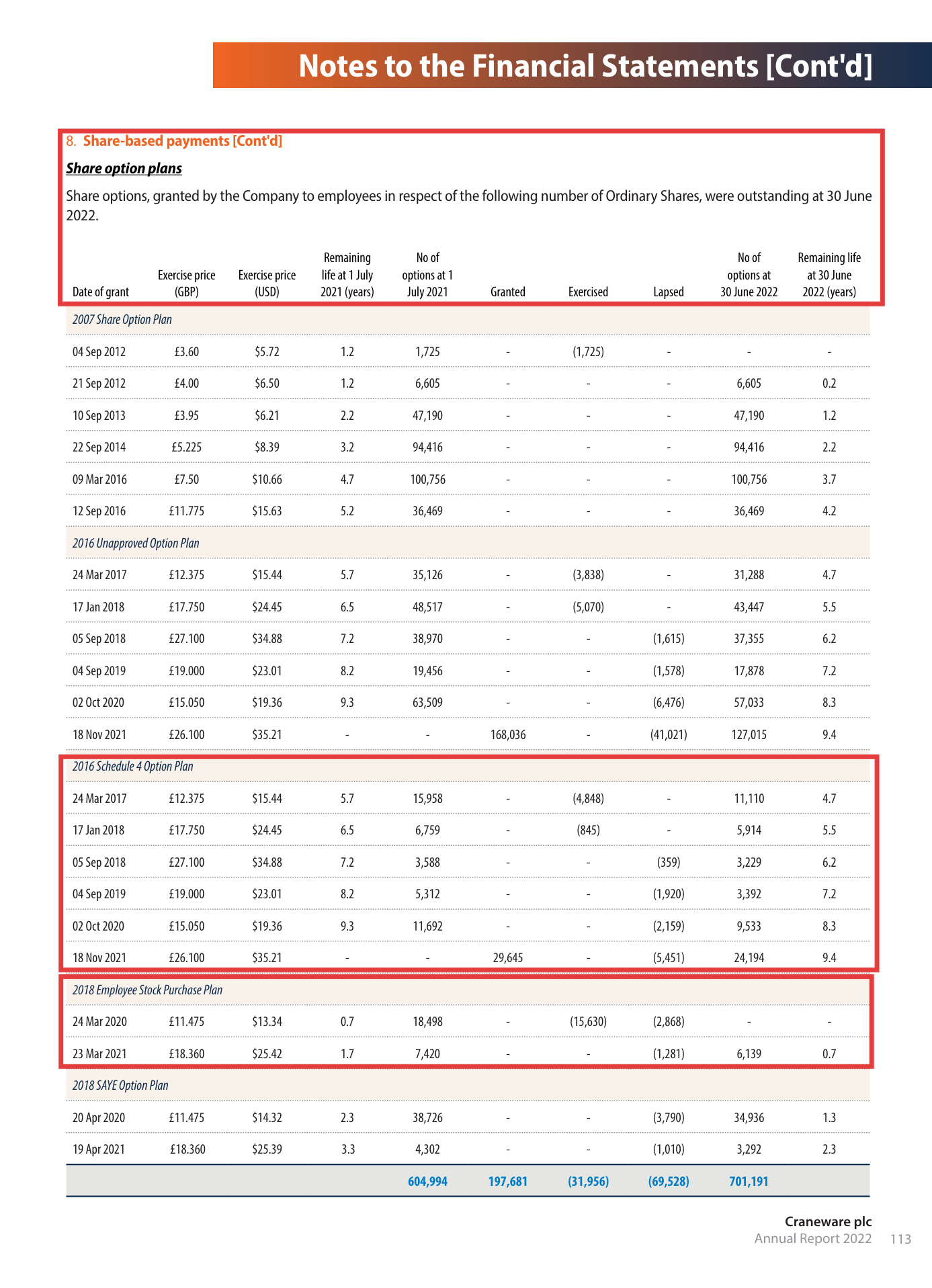} \\
    \bottomrule
\end{tabular}
\caption{QA Dataset Example 11: An Example of Table-Comparison and Sorting Question}
\label{tab:table-compare}
\end{table*}

\begin{table*}[h!]
\centering
\begin{tabular}{lp{10cm}} 
    \toprule
    \textbf{Query:} &In the 'Related party transactions' of the Craneware plc Annual Report and Financial Statements 2023, compare the share-based payments for Executive Directors and Other key management for 2023. Which category received higher payments?\\
    \textbf{Category:} &Table-Comparison and Sorting \\
    \textbf{Answer:} &For the year 2023, Executive Directors received \$929,609 in share-based payments, while Other key management received \$824,662. Executive Directors received higher payments.\\
    \textbf{Reference Image:} & \\
    &\includegraphics[width=0.6\textwidth]{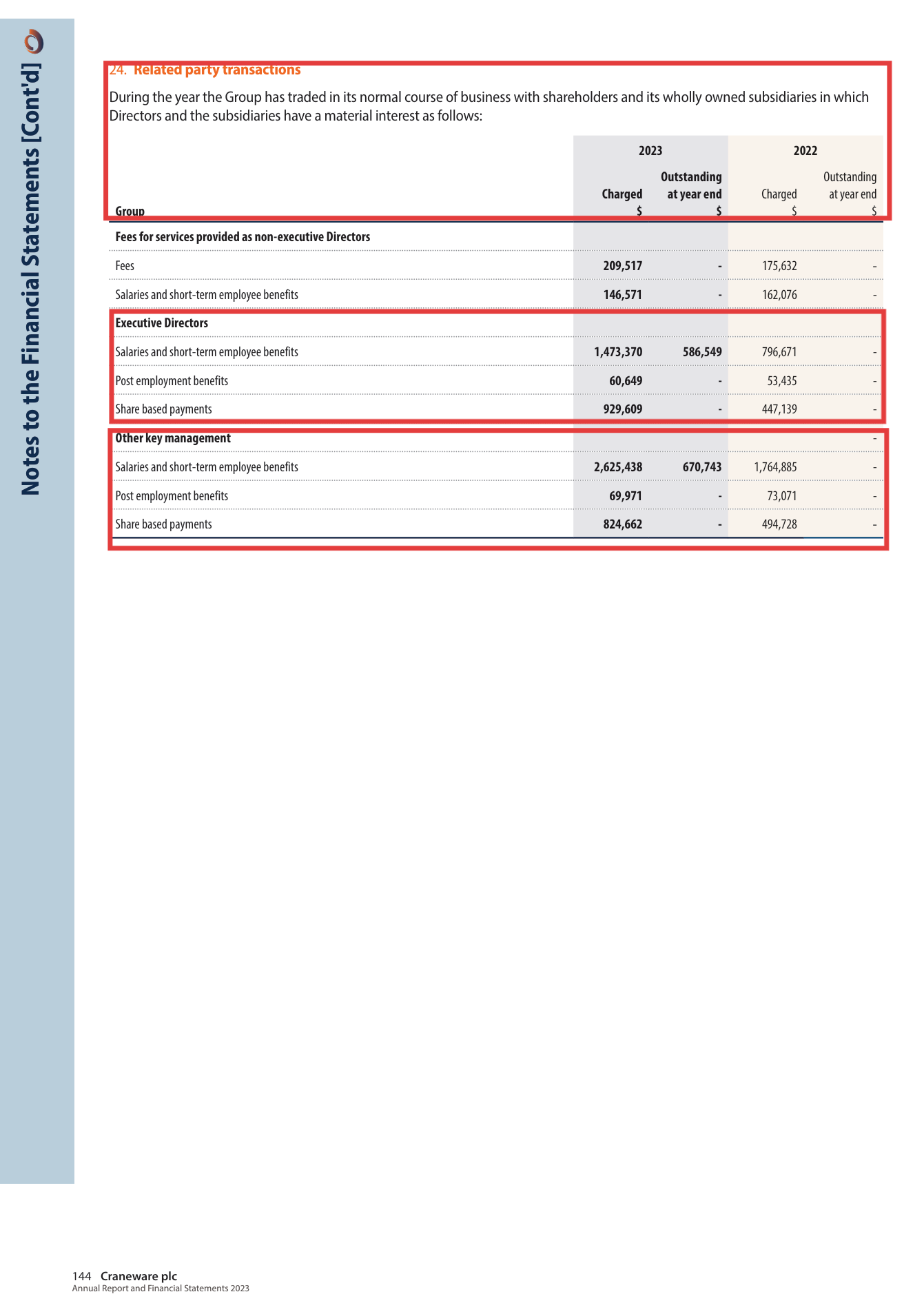} \\
    \bottomrule
\end{tabular}
\caption{QA Dataset Example 12: An Example of Table-Comparison and Sorting Question}
\label{tab:table-compare2}
\end{table*}

\begin{table*}[h!]
\centering
\begin{tabular}{lp{10cm}} 
    \toprule
    \textbf{Query:} & According to Ambac Financial Group, Inc' 2023 Form 10-K, during the years 2021 to 2023, which year had the highest Net premiums earned under Legacy Financial Guarantee Insurance?\\
    \textbf{Category:} &Multi-page\\
    \textbf{Answer:} &During the years 2021 to 2023, the highest net premiums earned by Legacy Financial Guarantee Insurance were in 2021, amounting to 46 million US dollars.\\
    \textbf{Reference Image:} & \\
    &\includegraphics[width=0.6\textwidth]{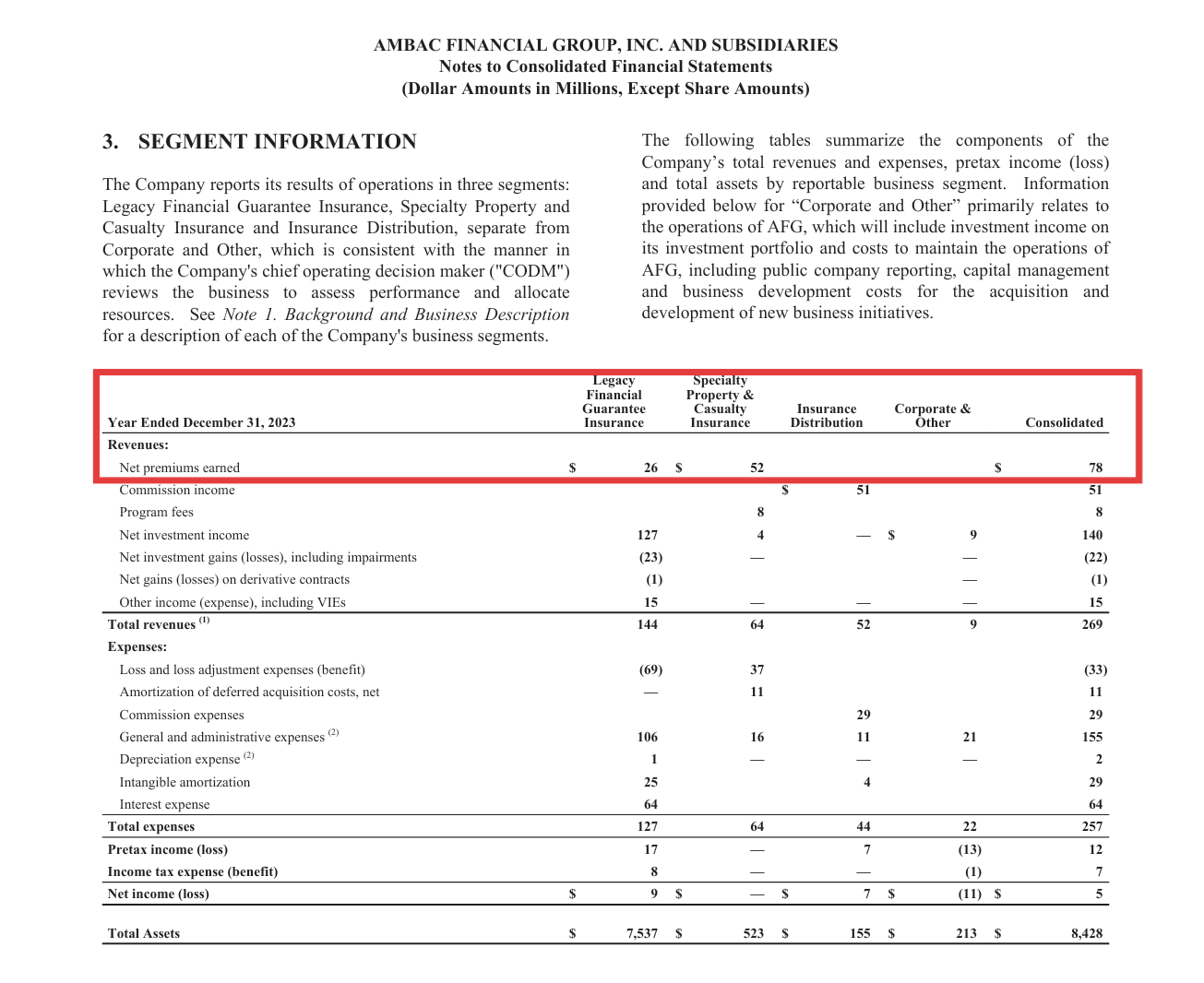}
    \includegraphics[width=0.6\textwidth]{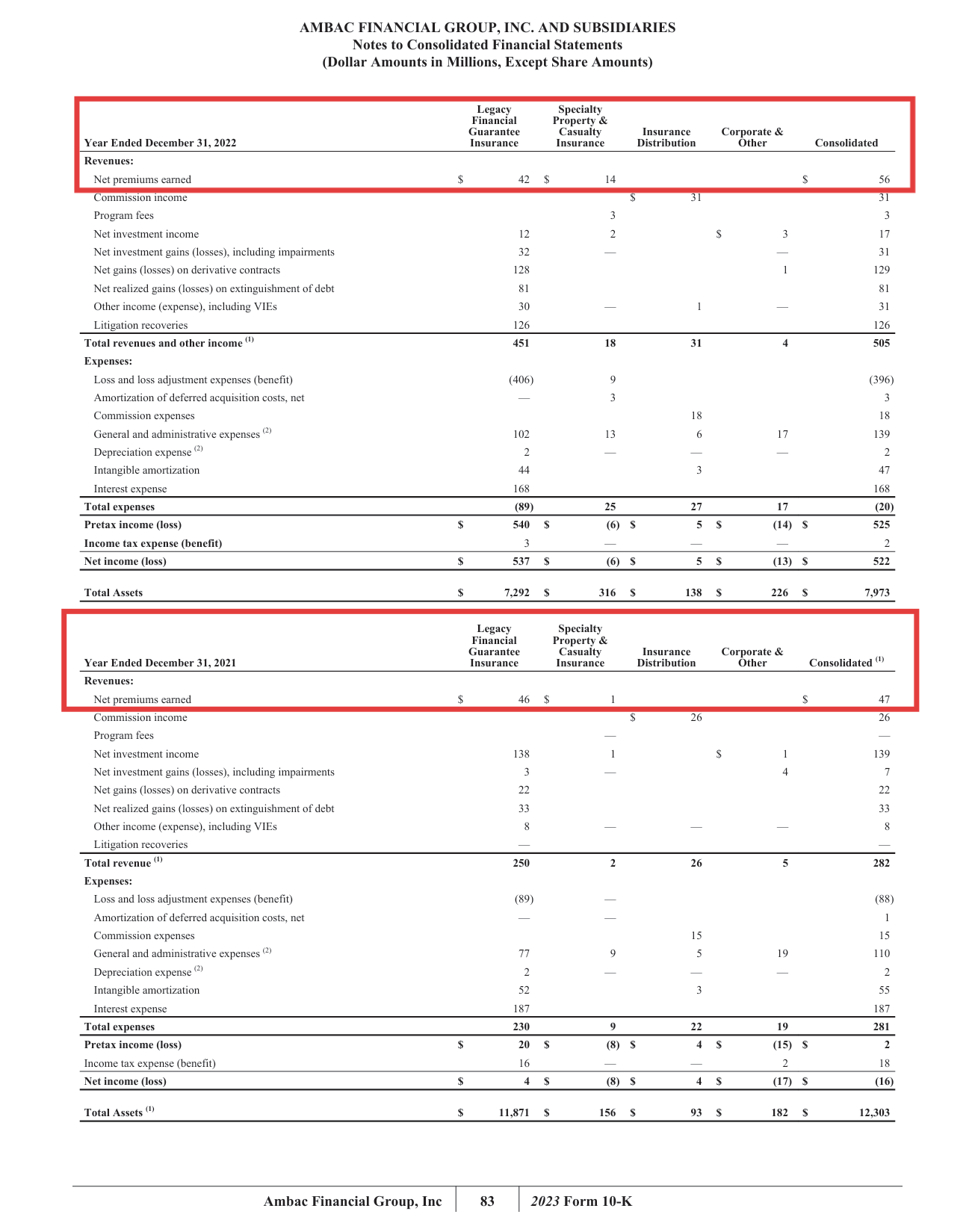} \\
    \bottomrule
\end{tabular}
\caption{QA Dataset Example 13: An Example of Multi-page Question}
\label{tab:multipage}
\end{table*}

\begin{table*}[h!]
\centering
\begin{tabular}{lp{10cm}}
    \toprule
    \textbf{Query:} & According to Ambac Financial Group, Inc.\ 2023 Form 10-K, how did the total value of Level-3 Financial Assets and Liabilities change for AMBAC Financial Group, Inc.\ and its subsidiaries for each end of period from 2021 to 2023??\\
    \textbf{Category:} &Multi-page\\
    \textbf{Answer:} &The total value of Level-3 Financial Assets and Liabilities for AMBAC Financial Group, Inc. and its subsidiaries at the end of each period from 2021 to 2023 changed as follows: At the end of December 31, 2021, the total value was \$6,199 million; At the end of December 31, 2022, the total value was \$3,762 million; At the end of December 31, 2023, the total value was \$3,848 million. This shows a decrease in the total value from 2021 to 2022, followed by a slight increase from 2022 to 2023.\\
    \textbf{Reference Image:} & \\
    &\includegraphics[width=0.4\textwidth]{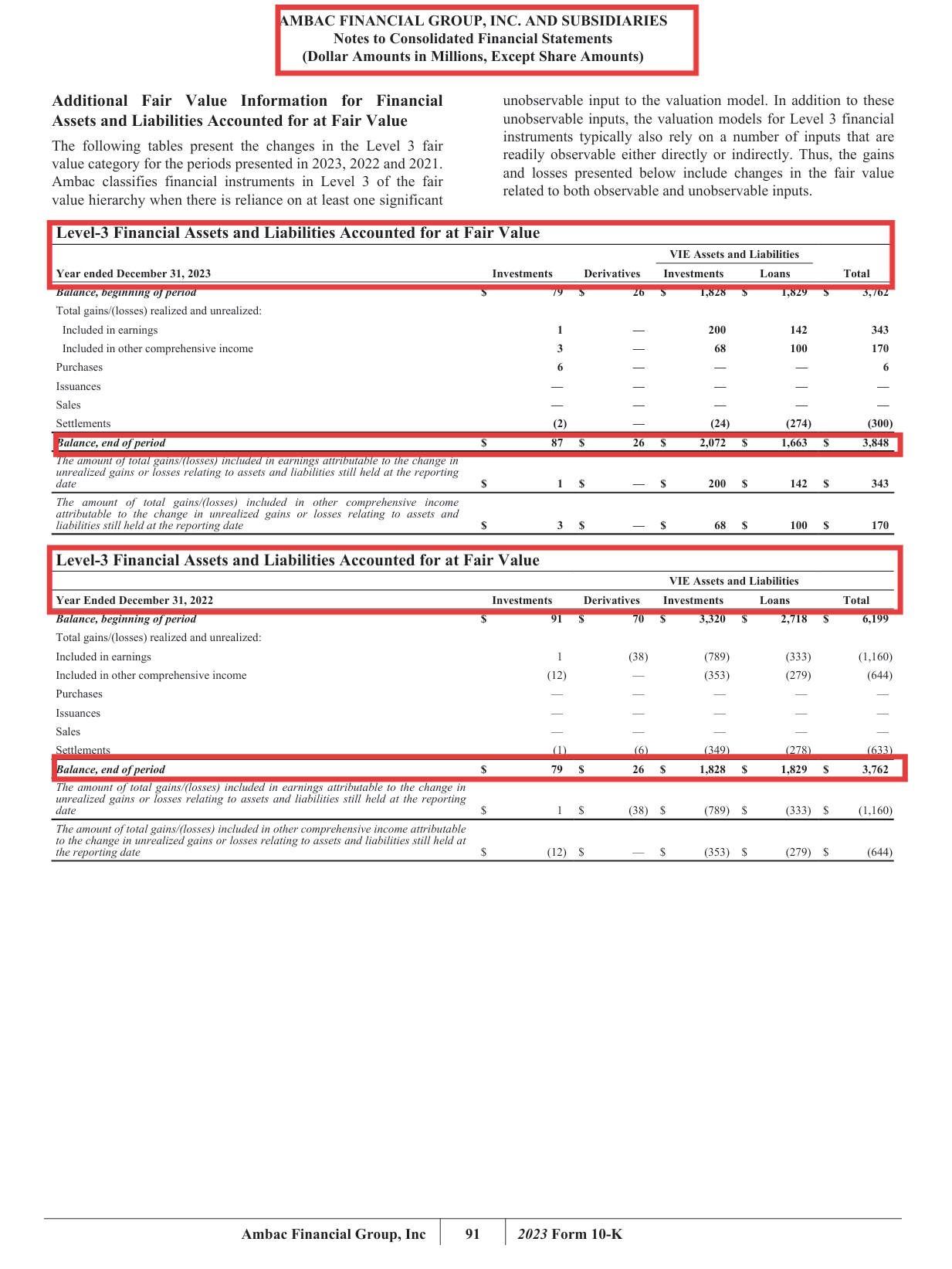}
    \includegraphics[width=0.4\textwidth]{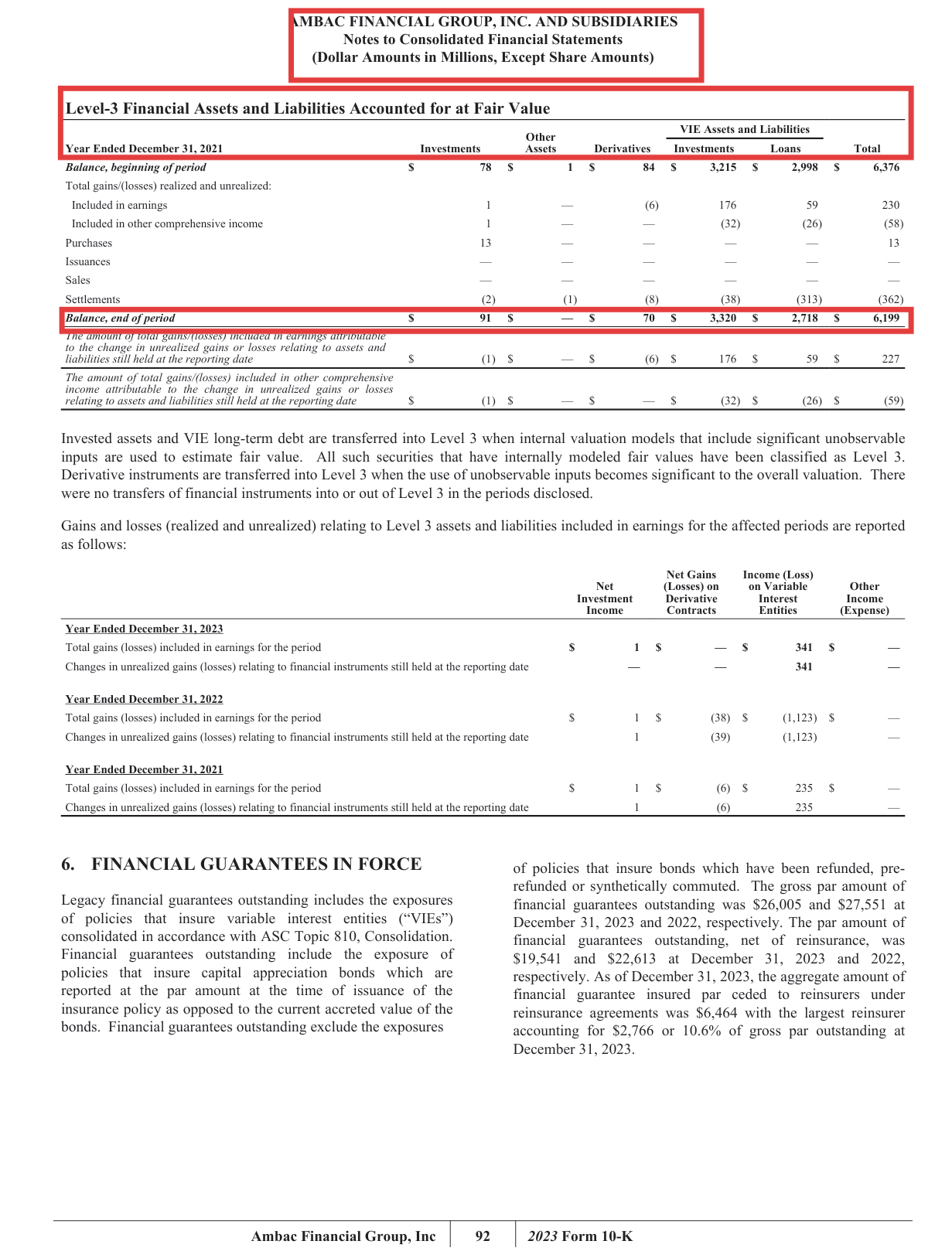} \\
    \bottomrule
\end{tabular}
\caption{QA Dataset Example 14: An Example of Multi-page Question}
\label{tab:multipage2}
\end{table*}

\clearpage
\begin{table*}[t]
\centering
\label{tab:qa_annotation_guideline}
\begin{tabular}{@{}p{0.95\textwidth}@{}}
\toprule
\textbf{Annotation Guideline for QA Pairs Verification} \\
\textbf{GUIDELINE:} Please verify the QA pairs produced by GPT-4o. For each sample, you may choose to retain, revise, or discard it. Your decision should be based on the following four criteria: \\

\textbf{Verification Criteria:} \\
\textbf{1. Query Clarity:} The query should be specific and unambiguous, targeting a particular topic in a document, and avoiding vague or overly general questions. \\

\textbf{2. Answer Correctness:} The answer must be factually correct and directly supported by the visual content. It should not include hallucinations or inferred information beyond what is presented. If calculation is involved, the answer should be accurate. \\

\textbf{3. Category Appropriateness:} The question should match its assigned category (e.g., table-numerical calculations). Mislabelled or ambiguous categories should lead to revision or rejection. \\

\textbf{4. Correctness of Multi-Page Sources:} For multi-page queries, if the answer is derived from multiple pages, all referenced page sources must be accurately identified. \\

\textbf{Decision Rule:} Retain the data if all criteria are met, revise it if there are minor issues (e.g., unclear query, incorrect category), and discard it if there are major errors or cannot be fixed reliably. \\

\bottomrule
\caption{Annotation guideline for QA Pairs Verification}
\label{tab:qa_annotation_guideline}
\end{tabular}
\end{table*}

\begin{table*}[t]
\centering
\label{tab:citation_annotation_guideline}
\begin{tabular}{@{}p{0.95\textwidth}@{}}
\toprule
\textbf{Annotation guideline for the Rating-based Human Evaluation} \\
\textbf{GUIDELINE:} Please evaluate the quality of the visual citation produced by the Retrieval-Augmented Generation system, rating it from score 0 to 5. Your rating should adhere to the following criteria: \\
\textbf{Scoring Criteria:} \\
\textbf{0:} Error image, or no reference/empty reference box.\\
\textbf{1:} Correct image, but selected the wrong area, containing no readable information or completely unrelated to the referenced content.\\
\textbf{2:} Correct image, area roughly related, but significantly offset, causing key information to be missing.\\
\textbf{3:} Correct image and roughly correct area, with offset or incomplete capture, information discernible but affecting reading experience.\\
\textbf{4:} Correct image and area, referenced information complete and accurate, with minor offset, or includes some redundant content (e.g., extra paragraphs, whitespace), but does not affect reading.\\
\textbf{5:} Perfect match. Image and area completely accurate, no offset, no redundancy, precise boundaries, referenced content clear and complete.\\
\bottomrule
\caption{Annotation guideline for the Rating-based Human Evaluation}
\label{tab:citation_annotation_guideline}
\end{tabular}
\end{table*}
\appendix
\end{document}